\documentclass[a4paper,fleqn]{cas-dc}

\usepackage[numbers]{natbib}
\usepackage{booktabs}
\usepackage{graphicx}

\usepackage{array}
\usepackage{longtable}
\usepackage{fancyhdr}
\usepackage{lastpage}

\def\tsc#1{\csdef{#1}{\textsc{\lowercase{#1}}\xspace}}
\tsc{WGM}
\tsc{QE}
\tsc{EP}
\tsc{PMS}
\tsc{BEC}
\tsc{DE}

\emergencystretch=\maxdimen
\begin{document}
\let\WriteBookmarks\relax
\def\floatpagepagefraction{1}
\def\textpagefraction{.001}
\shorttitle{A Two-Stage Deep Ensemble Framework for Brain-Behavior Modeling}
\shortauthors{Rahman et~al.}

\title [mode = title]{\textbf{Brain-Behavior Modeling from Multimodal MRI using Opportunistic Pretraining and Common-Cohort Adaptation}}

\author[1,2]{Md. Tanvir Rahman} [orcid=0000-0002-6173-0483]
\cormark[1]
\ead{tanvirrahman@mbstu.ac.bd}
\ead{m.t.rahman@student.uq.edu.au}
\credit{Conceptualization, Methodology, Formal analysis, Investigation, Data Curation, Resources, Visualization, Writing - Original Draft}

\author[3]{Taslima Ferdaus Shuva}[orcid=0000-0002-4672-4528]
\credit{Visualization, Writing - Original Draft}

\author[4]{Nabil Anan Orka} [orcid=0000-0001-5251-2137]
\credit{Methodology, Formal analysis}

\author[1]{Asaduzzaman Khan}[orcid=0000-0003-4188-2065]
\credit{Supervision, Writing - Review \& Editing}

\author[1,5,6]{Mohammad Ali Moni} [orcid=0000-0003-0756-1006]
\cormark[2]
\ead{m.moni@uq.edu.au}
\ead{mmoni@csu.edu.au}
\credit{Validation, Supervision, Writing - Review \& Editing}

\affiliation[1]{organization={School of Health and Rehabilitation Sciences, The University of Queensland},
            state={QLD},
            postcode={4072}, 
            country={Australia}}

\affiliation[2]{organization={Department of Information and Communication Technology, Mawlana Bhashani Science and Technology University},
            city={Tangail},
            postcode={1902}, 
            country={Bangladesh}}

\affiliation[3]{organization={Department of Computer Science and Engineering, Daffodil International University},
            addressline={Daffodil Smart City (DSC), Birulia, Savar}, 
            city={Dhaka},
            postcode={1216}, 
            country={Bangladesh}}

\affiliation[4]{organization={Graduate School of Biomedical Sciences, Icahn School of Medicine at Mount Sinai},
            city={New York},
            postcode={10029},
            state={NY},
            country={USA}}

\affiliation[5]{organization={AI and Cyber Futures Institute, Charles Sturt University},
            city={Bathurst},
            postcode={2795}, 
            state={New South Wales},
            country={Australia}}

\affiliation[6]{organization={Rural Health Research Institute, Charles Sturt University},
            city={Orange},
            postcode={2800}, 
            state={New South Wales},
            country={Australia}}

\begin{abstract}
Incomplete imaging coverage limits sample sizes for multimodal cognitive prediction. We evaluated a two-stage opportunistic deep ensemble using baseline Adolescent Brain Cognitive Development Study data for cross-sectional prediction of fluid, crystallized, and total cognition. Separate neural models processed structural/diffusion, resting-state functional, and task-based functional magnetic resonance imaging (MRI) data, with demographic and acquisition-related covariates. Models were pretrained on fold-specific subsets of eligible modality cohorts (up to 9,686 participants), adapted to common training cohorts, and fused using Ridge regression; the trimodal cohort included 4,216 participants. Five-fold family-aware cross-validation compared complete training protocols and fusion against matched constituent and covariate-only models. Two-stage training increased mean test coefficient of determination (R²) in 24 of 27 comparisons, with a median absolute gain of 0.0081 (range: -0.0018 to 0.0259). Fusion increased mean R² by 0.0051–0.0333 over the strongest constituent and reduced mean absolute error by 0.0067–0.0976 score units. Task-plus-resting-state and trimodal ensembles reduced absolute error across all targets in family-aware permutation tests (nominal $p\leq0.0108$). Task-plus-resting-state fusion achieved the highest observed mean R² for fluid (0.2222) and total cognition (0.3665), while structural-plus-resting-state fusion led for crystallized cognition (0.3706). Corresponding R² gains over covariate-only references were 0.0805, 0.1180, and 0.0890, respectively. Ablations suggested expanded pretraining accounted for much of the improvement. Gradient-based attribution identified target-dependent input patterns emphasized by the fitted models. These findings support using partially observed cohorts for model development, with modest fusion gains. If externally and prospectively validated across patient populations and acquisition settings, this approach may support MRI-informed cognitive phenotyping.
\end{abstract}


\maketitle

\section{Introduction}
\label{sec:introduction}
Understanding individual differences in cognitive ability from MRI requires models that integrate information across distributed brain systems. Structural and diffusion measurements characterize regional anatomy and tissue properties, whereas resting-state and task-based functional MRI provide complementary descriptions of intrinsic connectivity and task-evoked activity \cite{bassett2017network,greene2018task}. Combining these measurements may improve prediction, but it also introduces a data availability problem: participants with usable data across all selected modalities may form a substantially smaller cohort than those eligible for each modality separately. Restricting all model development to this intersection leaves potentially informative observations unused.

The value of multimodal integration must nevertheless be established empirically. A systematic review of deep learning for neurocognition identified variation in modality use, evaluation procedures, and interpretability practices, limiting direct comparison across studies \cite{rahman2026understanding}. Comparative evidence also indicates that the predictive value of an imaging modality depends on the other measurements available and the outcome being modeled \cite{ooi2022comparison}. In addition, prediction of between-person cognitive differences should be distinguished from prediction of within-person cognitive change \cite{konopkina2026multimodal}. These considerations motivate evaluating both the training strategy and the incremental value of fusion under clearly defined partitions of participants.

Existing approaches include multimodal graph convolutional networks \cite{qu2021ensemble} and opportunistic prediction-level stacking, which has been applied to brain-age estimation and cognitive prediction using participants with incomplete modality coverage \cite{engemann2020combining,pat2022longitudinally}. Modality-specific deep predictors have also been developed for adolescent structural MRI and resting-state functional connectomes \cite{saha2021predicting,thapaliya2025brain}. In the ABCD cohort, opportunistic learning has combined modality-specific elastic-net predictors with a random-forest meta-model \cite{pat2022longitudinally}. Other incomplete-modality approaches learn representations from available inputs, impute missing neuroimages, or preserve modality-specific information during fusion \cite{havaei2016hemis,zhang2022mmformer,liu2022assessing,xu2025domain}, while meta-matching transfers predictive information from larger datasets to new phenotypes \cite{he2022metamatching}. A recent scoping review highlights incomplete data and heterogeneous fusion strategies as continuing challenges in multimodal medical AI \cite{schouten2025navigating}. These approaches set precedents for using partially observed cohorts and transferring learned information, but they do not identify the most effective training and adaptation strategy for a given set of heterogeneous deep predictors.

The present study addresses a specific design question: Can modality-specific deep models, trained on expanded participant pools, be adapted to a common training cohort and effectively combined at the prediction level? Differences in acquisition requirements and quality-control criteria result in partially overlapping modality-eligible cohorts \cite{casey2018adolescent,hagler2019image}, which may therefore have different participant distributions. Adapting the models to a common cohort allows updating each predictor using the population on which the fusion is developed while retaining information learned from additional modality-eligible participants. This approach must accommodate differences between graph-based and dense architectures without assuming a shared latent representation.

Evaluation of this strategy also requires careful separation of model development from testing. Leakage through preprocessing, feature selection, or related participants can inflate neuroimaging prediction estimates \cite{rosenblatt2024leakage}. Demographic and acquisition-related predictors can substantially contribute to performance, and their inclusion does not, by itself, eliminate confounding \cite{snoek2019how}. Accordingly, prediction gains should be assessed against covariate-only benchmarks and individual imaging predictors on matched test participants. Attribution analyses can further characterize which inputs the fitted models emphasize, but cannot establish biological validity or causal mechanisms solely from predictive importance.

We therefore evaluate a two-stage opportunistic deep ensemble for fluid, crystallized, and total cognition using baseline ABCD data. IntelliGNN models structural and diffusion-derived measurements \cite{rahman2026intellignn}, SPRN models resting-state connectivity and regional signal variance \cite{rahman2026sprn}, and MAC-Net models task-evoked activations and contrasts \cite{rahman2026macnet}. Each predictor is trained independently on its fold-specific modality-eligible pool after excluding held-out participants and their families. The resulting models undergo common-cohort adaptation with validation-based checkpoint selection, and a Ridge meta-regressor combines their predictions. The framework addresses incomplete modality coverage during training; each evaluated ensemble requires the modalities specified in its configuration at inference time.

Four questions guide the evaluation: (i) whether the complete opportunistic two-stage strategy improves prediction over common-cohort-only training; (ii) whether the defined ablation variants clarify the relative roles of pretraining, adaptation, freezing, and proximal regularization; (iii) whether prediction-level fusion improves performance over the strongest constituent and covariate-only benchmark within each configuration; and (iv) which imaging features and covariates are emphasized by the fitted predictors, and which attribution patterns recur across targets and modality configurations.

\begin{table*}[htpb]
\centering
\caption{Participant demographics, MRI characteristics, and cognitive scores following fold-based imputation.}
\label{tab:demographics_imputed}
\renewcommand{\arraystretch}{1.15}
\setlength{\tabcolsep}{8pt}
\resizebox{\textwidth}{!}{
\begin{tabular}{l ccccccc}
\toprule
\textbf{Characteristic} & \textbf{Struct} & \textbf{Task} & \textbf{Rest} & \textbf{Struct+Task} & \textbf{Struct+Rest} & \textbf{Task+Rest} & \textbf{All Modalities} \\
\midrule
\multicolumn{8}{l}{\textbf{Sample Information}} \\
\hspace{3mm} Subjects ($n$) & 9,686 & 7,287 & 5,555 & 6,845 & 5,344 & 4,356 & 4,216 \\
\hspace{3mm} Families ($n$) & 8,238 & 6,405 & 4,901 & 6,015 & 4,712 & 3,921 & 3,789 \\
\addlinespace
\multicolumn{8}{l}{\textbf{Demographics \& Anthropometrics}} \\
\hspace{3mm} Age (months) & 119.1 $\pm$ 7.5 & 119.5 $\pm$ 7.5 & 119.8 $\pm$ 7.5 & 119.6 $\pm$ 7.5 & 119.8 $\pm$ 7.5 & 120.1 $\pm$ 7.5 & 120.1 $\pm$ 7.5 \\
\hspace{3mm} BMI & 18.7 $\pm$ 4.1 & 18.6 $\pm$ 3.9 & 18.3 $\pm$ 3.7 & 18.5 $\pm$ 3.9 & 18.3 $\pm$ 3.7 & 18.2 $\pm$ 3.7 & 18.2 $\pm$ 3.6 \\
\hspace{3mm} Sex \\
\hspace{6mm} Male & 4,970 (51\%) & 3,706 (51\%) & 2,692 (48\%) & 3,461 (51\%) & 2,587 (48\%) & 2,100 (48\%) & 2,032 (48\%) \\
\hspace{6mm} Female & 4,716 (49\%) & 3,581 (49\%) & 2,863 (52\%) & 3,384 (49\%) & 2,757 (52\%) & 2,256 (52\%) & 2,184 (52\%) \\
\addlinespace
\multicolumn{8}{l}{\textbf{Socioeconomic Status}} \\
\hspace{3mm} Parental Education \\
\hspace{6mm} $\le$ High School & 1,569 (16\%) & 975 (13\%) & 743 (13\%) & 895 (13\%) & 697 (13\%) & 492 (11\%) & 465 (11\%) \\
\hspace{6mm} Some College & 2,817 (29\%) & 2,001 (27\%) & 1,629 (29\%) & 1,873 (27\%) & 1,566 (29\%) & 1,198 (28\%) & 1,160 (28\%) \\
\hspace{6mm} Bachelor's & 2,774 (29\%) & 2,266 (31\%) & 1,704 (31\%) & 2,135 (31\%) & 1,648 (31\%) & 1,426 (33\%) & 1,385 (33\%) \\
\hspace{6mm} Postgraduate & 2,526 (26\%) & 2,045 (28\%) & 1,479 (27\%) & 1,942 (28\%) & 1,433 (27\%) & 1,240 (28\%) & 1,206 (29\%) \\
\hspace{3mm} Family Income \\
\hspace{6mm} $<$ \$50k & 2,507 (26\%) & 1,640 (23\%) & 1,295 (23\%) & 1,512 (22\%) & 1,222 (23\%) & 895 (21\%) & 851 (20\%) \\
\hspace{6mm} \$50k -- \$100k & 3,303 (34\%) & 2,477 (34\%) & 1,951 (35\%) & 2,325 (34\%) & 1,880 (35\%) & 1,522 (35\%) & 1,477 (35\%) \\
\hspace{6mm} $\ge$ \$100k & 3,876 (40\%) & 3,170 (44\%) & 2,309 (42\%) & 3,008 (44\%) & 2,242 (42\%) & 1,939 (45\%) & 1,888 (45\%) \\
\addlinespace
\multicolumn{8}{l}{\textbf{MRI Information \& Quality}} \\
\hspace{3mm} Manufacturer \\
\hspace{6mm} Siemens & 6,392 (66\%) & 4,866 (67\%) & 4,028 (73\%) & 4,674 (68\%) & 3,942 (74\%) & 3,184 (73\%) & 3,131 (74\%) \\
\hspace{6mm} GE & 2,141 (22\%) & 1,674 (23\%) & 1,114 (20\%) & 1,488 (22\%) & 1,015 (19\%) & 868 (20\%) & 792 (19\%) \\
\hspace{6mm} Philips & 1,153 (12\%) & 747 (10\%) & 413 (7\%) & 683 (10\%) & 387 (7\%) & 304 (7\%) & 293 (7\%) \\
\hspace{3mm} Mean Motion (FD) \\
\hspace{6mm} Structural (dMRI) & 1.3 $\pm$ 0.4 & - & - & 1.3 $\pm$ 0.3 & 1.2 $\pm$ 0.3 & - & 1.2 $\pm$ 0.3 \\
\hspace{6mm} Task (N-Back) & - & 0.4 $\pm$ 0.4 & - & 0.4 $\pm$ 0.4 & - & 0.3 $\pm$ 0.3 & 0.3 $\pm$ 0.3 \\
\hspace{6mm} Task (MID) & - & 0.4 $\pm$ 0.3 & - & 0.4 $\pm$ 0.3 & - & 0.3 $\pm$ 0.2 & 0.3 $\pm$ 0.2 \\
\hspace{6mm} Functional Rest & - & - & 0.2 $\pm$ 0.1 & - & 0.2 $\pm$ 0.1 & 0.2 $\pm$ 0.1 & 0.2 $\pm$ 0.1 \\
\addlinespace
\multicolumn{8}{l}{\textbf{Intelligence Targets}} \\
\hspace{3mm} $G_f$ & 92.0 $\pm$ 10.4 & 93.6 $\pm$ 9.6 & 93.0 $\pm$ 10.1 & 93.7 $\pm$ 9.5 & 93.1 $\pm$ 10.0 & 94.3 $\pm$ 9.4 & 94.3 $\pm$ 9.4 \\
\hspace{3mm} $G_c$ & 86.7 $\pm$ 7.0 & 87.6 $\pm$ 6.6 & 87.2 $\pm$ 6.7 & 87.7 $\pm$ 6.5 & 87.2 $\pm$ 6.6 & 87.9 $\pm$ 6.3 & 87.9 $\pm$ 6.3 \\
\hspace{3mm} $G_t$ & 86.7 $\pm$ 8.9 & 88.2 $\pm$ 8.1 & 87.6 $\pm$ 8.5 & 88.3 $\pm$ 8.1 & 87.6 $\pm$ 8.5 & 88.8 $\pm$ 7.9 & 88.8 $\pm$ 7.8 \\
\bottomrule
\multicolumn{8}{p{1.2\linewidth}}{\footnotesize \textit{Note:} $G_c$: Crystallized Intelligence; $G_f$: Fluid Intelligence; $G_t$: Total Intelligence. Continuous variables are reported as mean $\pm$ standard deviation, and categorical variables are reported as frequency (\%). Missing values for BMI, Parental Education, and Family Income were imputed independently within each fold of the 5-fold cross-validation scheme, using the respective training-set medians to prevent information leakage. Hyphens ($-$) indicate metrics not applicable to the specified scan configuration.}
\end{tabular}
}
\end{table*}

\begin{figure*}[htpb]
    \centering
    \includegraphics[width=\textwidth]{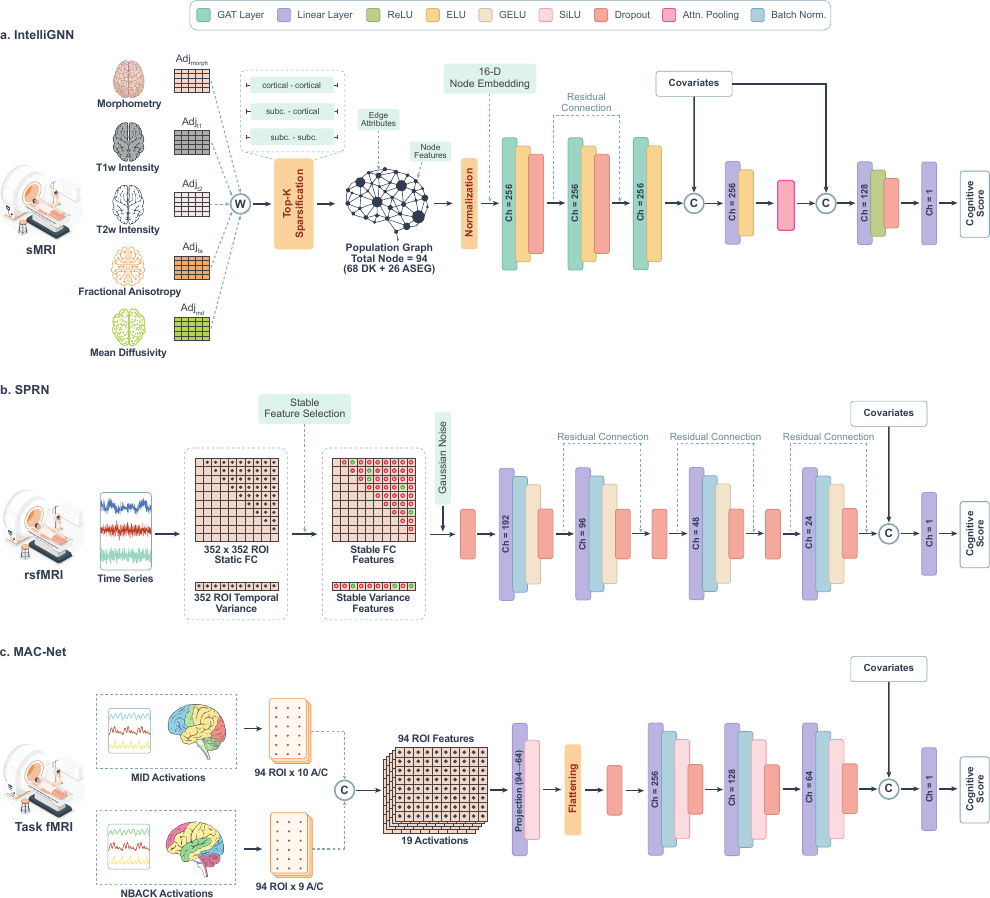}
    \caption{Modality-specific architectures for cognitive prediction. (a) IntelliGNN processes a population-level graph with 94 regional nodes, target-specific structural and diffusion feature masks, edge attributes, and learned node embeddings. Covariates are incorporated before attention pooling and again at the regression head. (b) SPRN processes selected resting-state connectivity and regional variance features through projected residual layers, followed by covariate concatenation at the output head. (c) MAC-Net processes regional N-back and MID measurements and concatenates covariates at the output head. Each branch is fitted separately for \(G_f\), \(G_c\), and \(G_t\) to predict a transformed cognitive target; outputs are inverse-transformed before score-scale evaluation and fusion. Models are trained independently and do not share a latent representation.}
    \label{fig:architectures}
\end{figure*}

\begin{figure*}[htpb]
    \centering
    \includegraphics[width=\textwidth]{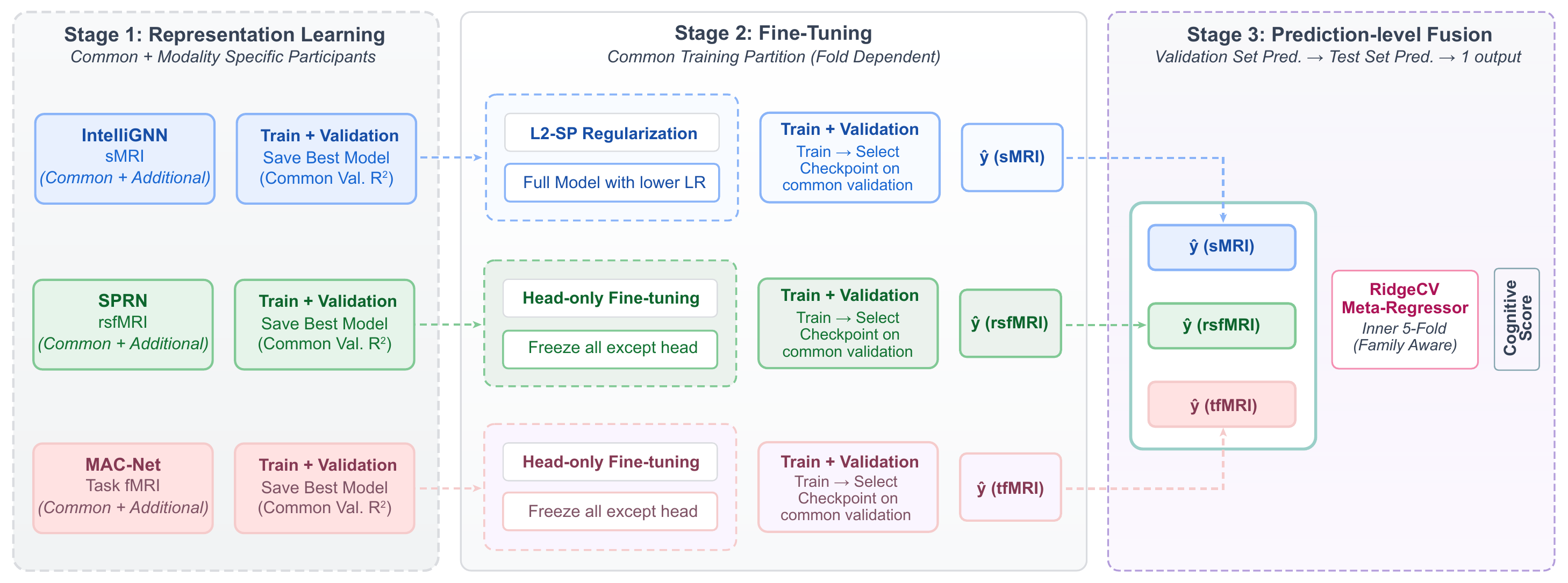}
    \caption{Overview of the dynamic multimodal late-fusion pipeline. Structural (sMRI), resting-state (rsfMRI), and task (tfMRI) modalities are processed independently using dedicated deep learning architectures (IntelliGNN, SPRN, and MAC-Net). Following modality-specific two-stage training, predictions for the common validation participants are extracted to calibrate the fusion weights. Prediction-level fusion is performed using a RidgeCV meta-regressor, which employs an internal 5-fold cross-validation strategy on the validation set. This internal loop selects the \(L_2\) regularization penalty while keeping family members together within each split. The validation partition is also used for neural checkpoint selection, so the fusion procedure constitutes validation-set blending. The dynamically regularized coefficients are then applied to the hold-out test predictions to generate the final unified cognitive prediction.}
    \label{fig:two_stage_fusion}
\end{figure*}

\section{Methodology}
\label{sec:methodology}
This study used three modality-specific neural architectures to model complementary MRI information. IntelliGNN \cite{rahman2026intellignn} processed structural and diffusion-derived regional measures (sMRI), SPRN \cite{rahman2026sprn} processed resting-state functional connectivity and regional signal variance (rsfMRI), and MAC-Net \cite{rahman2026macnet} processed task-evoked regional activation measures from the N-back and Monetary Incentive Delay tasks (tfMRI). The three models were trained independently because their input structures and learning procedures differed; their predictions were combined only at the final prediction-fusion stage. The following subsections describe the study cohorts, feature representations, evaluation framework, model architectures, and multimodal fusion procedure.

\subsection{Study Design, Participants, and Cognitive Outcomes}
We developed a modality-specific prediction framework in which structural (sMRI), resting-state (rsfMRI), and task-based (tfMRI) functional MRI models were trained independently, adapted to participants with complete data for a specified combination of modalities, and combined at the prediction level. The sMRI branch incorporated both structural and diffusion-derived regional measurements.

Data were obtained from the Adolescent Brain Cognitive Development (ABCD) Study Tabulated Data Release 5.1 and resting-state connectivity derivatives from the ABCD-HCP processing pipeline \cite{casey2018adolescent,hagler2019image}. Analyses used baseline assessments only. Site 22 was excluded from all downstream analyses due to its small baseline sample size (\(n=36\)), leaving 21 sites in the analysis. Official ABCD documentation notes that Site 22 was discontinued after baseline and that most participants transferred to other sites; it also permits retaining all 22-site labels for baseline-only analyses \cite{abcdsitedocumentation}. The exclusion was therefore an analysis-specific decision rather than a mandatory ABCD exclusion. Of 11,868 merged participants, we excluded Site 22 (\(n=36\)), missing sex (\(n=3\)), missing or out-of-range age (\(n=1\); 90--150 months), and missing cognitive outcomes (\(n=245\)), leaving 11,583 participants before modality-specific eligibility filtering. Outcomes were the NIH Toolbox uncorrected Fluid (\(G_f\)), Crystallized (\(G_c\)), and Total (\(G_t\)) Cognition Composite scores \cite{weintraub2013cognition}. They were modeled as continuous outcomes and not interpreted as diagnostic thresholds.

\subsection{Modality Eligibility and Descriptive Characteristics}
sMRI eligibility required complete structural and diffusion features, ABCD inclusion flags for \(T_1\)- and \(T_2\)-weighted MRI and diffusion MRI, and available intracranial volume and diffusion-motion measures. Participants with a FreeSurfer manual-QC code of 0 were excluded; a missing manual-QC entry was permitted when the remaining criteria were met. tfMRI eligibility required complete N-back and Monetary Incentive Delay (MID) features, inclusion flags for both tasks, and available motion measures for both tasks. rsfMRI eligibility required an extracted connectivity record, an inclusion flag, and the availability of resting-state motion. No additional numerical motion threshold was applied during cohort assembly. The eligible unimodal cohorts comprised 9,686 sMRI, 7,287 tfMRI, and 5,555 rsfMRI participants. Pairwise intersections comprised 6,845 sMRI+tfMRI, 5,344 sMRI+rsfMRI, and 4,356 tfMRI+rsfMRI participants; the trimodal intersection comprised 4,216 participants. These overlapping cohorts defined the common populations for the corresponding configurations.

Participant characteristics included age, sex at birth, highest parental education, combined household income, body mass index (BMI), family identifier, study site, scanner device serial number, intracranial volume, and modality-specific mean-motion measures. Numeric sentinel codes were converted to missing values. BMI was derived from recorded weight and height and set to missing when outside the prespecified plausible range. Descriptive statistics were generated separately for each analytic cohort (Table \ref{tab:demographics_imputed}). BMI, parental education, and household income were imputed for each outer-test participant using medians calculated from the corresponding training participants in that cohort. The five disjoint test partitions were then combined so that each participant contributed once. Continuous variables are reported as mean (SD) and categorical variables as count (\%); family counts refer to distinct non-missing family identifiers. This descriptive procedure was separate from model-specific training imputation. Before descriptive imputation, the trimodal cohort had 9 missing BMI values, 3 missing education values, and 271 missing income values.

\subsection{Neuroimaging Feature Extraction}
Regional structural and task measures were extracted from baseline derivative tables, retaining the first record for duplicated imaging entries. Within each branch, participant sets were intersected across selected tables to require complete regional measurements. Arrays were stored in a consistent participant--region order. We assessed missingness and extreme standardized values but did not remove observations or modify saved feature values on this basis.

\subsubsection{Structural and Diffusion MRI}
The sMRI representation comprised 68 Desikan--Killiany cortical and 26 bilateral FreeSurfer automated-segmentation (ASEG) regions (Supplementary Table S1). Sixteen measurements were extracted per cortical region: thickness, area, volume, and sulcal depth; gray-matter, white-matter, and gray--white contrast measures for both \(T_1\)- and \(T_2\)-weighted images; and fractional anisotropy (FA) and mean diffusivity (MD) measures for gray matter, white matter, and gray--white contrast. Each ASEG region contributed volume, \(T_1\) intensity, \(T_2\) intensity, FA, and MD. This yielded \(68\times16+26\times5=1,218\) measured values per participant, saved as 21 measurement arrays. For graph modeling, these values were embedded in a \(94\times21\) tensor; cortical measurements occupied the first 16 channels and ASEG measurements the final five channels, with unused positions zero-filled.

\subsubsection{Task-based Functional MRI}
Task features comprised nine N-back conditions or contrasts and ten MID contrasts, mapped to the same 94 regions (Supplementary Table S2) and to the same 68 cortical and 26 ASEG regional orderings, yielding 1,786 values per participant. ASEG entries outside the specified regional ordering were excluded to align the task tables with the selected 26-region layout.

\subsubsection{Resting-State Functional MRI}
Baseline rsfMRI data comprised the strict upper triangle of each \(352\times352\) connectivity matrix (61,776 values) and 352 regional-variance measures based on the Gordon 333 and ASEG parcellations (Supplementary Table S1). Connectivity values were complete, whereas variance values were handled by model-specific imputation. Matrix entries were used as supplied, avoiding additional connectivity transformations during extraction.

\subsection{Family-Aware Multimodal Cross-Validation Framework}
Let \(S_m\) denote the eligible cohort for modality \(m\). For modality configuration \(\mathcal{M}\), the common and additional participant sets were defined as

\begin{equation}
C_{\mathcal{M}}=\bigcap_{m\in\mathcal{M}}S_m,
\qquad
E_{\mathcal{M}}=\left(\bigcup_{m\in\mathcal{M}}S_m\right)\setminus C_{\mathcal{M}}.
\end{equation}

Only common participants were eligible for validation and testing. Additional participants were strictly eligible for training using available modality-specific imaging data. Common participants were partitioned by five-fold stratified group cross-validation, with recorded family identifiers defining groups and missing identifiers replaced by participant identifiers to create singleton groups. Within each outer training-plus-validation partition, a family-aware inner split defined common training and validation subsets, yielding approximate common-cohort proportions of 64\% training, 16\% validation, and 20\% testing. Additional participants were excluded from Stage 1 when their recorded family occurred in the common validation or test partition. Partitions were generated independently for each modality configuration. Thus, comparisons within a configuration used identical test participants, whereas comparisons across configurations were descriptive because eligible cohorts and training pools differed.

\subsection{Covariates and Training-Partition Preprocessing}
All three architectures included age, intracranial volume, highest parental education, combined household income, BMI, recorded sex, and scanner device serial number. MAC-Net additionally included mean motion for the N-back and MID tasks, and SPRN included resting-state mean motion. IntelliGNN included dMRI mean motion for \(G_f\) and \(G_t\). These variables provide demographic, socioeconomic, and acquisition context \cite{hagler2019image}.

Numeric covariates were standardized using only training-partition means and standard deviations. Sex and scanner serial number were cast as strings and one-hot encoded using a fixed category vocabulary constructed from the configuration cohort. This preserved column identities across folds, including folds in which a category was absent from the training partition. Vocabulary construction used covariate values without cognitive outcomes, whereas numeric imputation and scaling were fitted to the designated training participants. The categorical vocabulary was therefore outcome-independent but was not estimated exclusively from each training fold. For single-stage baselines, preprocessing parameters were estimated from common training participants. For two-stage models, numeric imputers, imaging and covariate scalers, and target transformations were estimated from each model's Stage-1 training pool and retained during adaptation; the categorical vocabulary was fixed as described above. Each target was non-linearly mapped to a normal distribution using a training-fitted quantile transformer. Networks minimized mean squared error (MSE) within this transformed space; predictions were inverse-transformed for score-scale evaluation.

\subsection{Structural Graph Construction}
For each configuration, target, and outer fold, a population-level graph was generated using the designated common and family-aware additional graph-construction training pool. Measurements were categorized into five groups: morphometry, \(T_1\) intensity, \(T_2\) intensity, FA, and MD. Non-zero entries were standardized per node-feature pair across the graph-construction training pool. For each participant and node, available channels within a specified group were averaged, and Pearson correlations between node summaries across participants generated a \(94\times94\) matrix \(R^{(g)}\). Diagonals and undefined correlations were set to zero.

For \(G_c\) and \(G_t\), group weights were proportional to the mean absolute off-diagonal correlation:
\begin{equation}
\begin{gathered}
a_g = \frac{1}{\binom{94}{2}} \sum_{i < j} |R^{(g)}_{ij}|, \\
w_g = \frac{a_g}{\sum_{h \in \mathcal{G}} a_h + 10^{-7}}, \\
A = \sum_{g \in \mathcal{G}} w_g R^{(g)}.
\end{gathered}
\end{equation}
\(G_f\) utilized equal weights uniformly across all five groups. \(G_c\) leveraged morphometry and both intensity groups; \(G_t\) leveraged morphometry, FA, and MD. The identical group selection determined the node-feature mask.

Edges were selected based on absolute fused association strength. Each cortical node selected its \(K_{cc}\) strongest cortical neighbors; each ASEG node selected \(K_{cs}\) cortical neighbors and \(K_{ss}\) alternative ASEG neighbors. The undirected union of these selections was represented in both directions. The settings were \((K_{cc},K_{cs},K_{ss})=(10,10,4)\) for \(G_f\) and \((5,5,2)\) for \(G_c\) and \(G_t\). Each retained edge carried five signed group-correlation channels, with inactive groups set to zero. The fused matrix dictated edge selection, while the attention mechanism utilized the group-correlation attributes. Self-connections were added programmatically when loading the graph into the network. The topology remained rigorously fixed across the two training stages.

\subsection{Resting-State Feature Selection}
Connectivity and regional variance were screened jointly, executing separately for each configuration, target, and fold. The selection pool combined common training participants with family-permitted additional participants belonging to the eligible rsfMRI cohort. Variance entries were median-imputed, and the combined features were identically standardized using this pool. Ridge regularization was selected via five-fold family-grouped cross-validation, optimizing to maximize mean Pearson correlation with the untransformed cognitive outcome across 20 candidate penalties spaced between 100 and \(10^6\). Imputation and scaling were fitted once to the feature-selection training pool before internal cross-validation and were not refitted within each internal split. Common validation and outer test participants were excluded from this pool. Thus, the internal tuning procedure reused preprocessing estimates across its splits, while the outer test evaluation remained separate from feature selection.

Using the selected penalty, 100 bootstrap samples were drawn at the family level with replacement, retaining all participant rows for each sampled family. Within bootstrap \(b\), a given feature was selected when:
\begin{equation}
|\beta_j^{(b)}| > \mu_b + 1.5\sigma_b,
\end{equation}
where \(\mu_b\) and \(\sigma_b\) represent the mean and standard deviation of absolute coefficients across all features. Selection frequencies were aggregated across bootstrap samples, and features present in at least 50\% were retained for SPRN. This functioned as a stability-based screening rule rather than an inferential significance threshold. The bootstrap seed sequence originated from seed 42. Masks were fixed during Stage-2 adaptation.

\subsection{Covariate-Only Baselines}
To contextualize predictions from regional imaging features, five covariate-only regressors were evaluated: Ridge, Lasso, Elastic Net, Random Forest, and a Multilayer Perceptron (MLP). Inputs spanned the demographic, socioeconomic, intracranial volume, scanner, and motion covariates, which were appropriately structured for each configuration and target. These models deliberately excluded regional imaging features while fully retaining imaging-derived intracranial volume and scalar motion metrics.

Training combined common and family-permitted additional participants across the specified configuration. Missing numeric covariates were populated using training medians; numeric scalers and a normal-output target quantile transformer were fitted entirely on training data. Categorical inputs referenced the common string-based vocabulary previously detailed. Hyperparameters were selected using a predefined split that exactly matched the saved common validation partition and minimized transformed-target MSE. The selected estimator was subsequently refitted on the unified training and validation data, strictly retaining the training-fitted preprocessing artifacts. Test predictions were inverse-transformed and evaluated against the original observed ground-truth outcomes.

Ridge penalties explored \(\{0.1,1,10,100,1000\}\) and Lasso penalties \(\{0.001,0.01,0.1,1,10\}\). Elastic Net integrated penalties \(\{0.01,0.1,1,10\}\) with mixing ratios \(\{0.2,0.5,0.8\}\). Random forests varied tree counts \(\{100,300\}\), maximum depths \(\{\mathrm{unrestricted},10,20\}\), and minimum leaf sizes \(\{1,4\}\). MLPs explored hidden widths \(\{(64,32),(128,64),(64)\}\), regularization strengths \(\{10^{-4},10^{-3},10^{-2}\}\), and initial learning rates \(\{0.001,0.01\}\), capped at 1,000 iterations incorporating internal early stopping. Lasso and Elastic Net configurations were capped at 5,000 iterations. Estimator seeds were fixed at 42.

\subsection{Modality-Specific Neural Architectures}
Figure \ref{fig:architectures} details the three specialized architectures: IntelliGNN, SPRN, and MAC-Net, informed by the corresponding modality-specific studies \cite{rahman2026intellignn,rahman2026sprn,rahman2026macnet}. The models had completely separate parameters and distinct training protocols; their predictions were combined only after each model was fitted.

\subsubsection{IntelliGNN (sMRI)}
Each structural node's 21-channel input was concatenated with a dedicated learned 16-dimensional node embedding. Three dense graph-attention layers of width 256 employed scaled dot-product attention equipped with edge-dependent biases. For an explicitly allowed edge \((i,j)\),
\begin{equation}
\begin{gathered}
s_{ij}=\frac{(W_qh_i)^\top(W_kh_j)}{\sqrt{256}}+w_e^\top e_{ij}+b_e,\\
h'_i=\operatorname{ELU}\left(\sum_j\operatorname{softmax}_j(s_{ij})W_vh_j\right).
\end{gathered}
\end{equation}
Non-edges were masked before executing the softmax function. A dropout of 0.4 was applied to the first and second layers, and the second-layer output received a residual addition from the initial layer. Following the third layer, covariates were concatenated directly to each node representation and projected to 256 dimensions utilizing ELU. A learned attention gate adaptively pooled these covariate-conditioned node representations. Covariates were concatenated anew to the pooled global vector, culminating in a two-layer regression MLP initialized with ReLU and a dropout of 0.4. Covariates, therefore, influenced both attention pooling and the final regression head; the pooled structural representation was not independent of the covariate inputs.

\subsubsection{Sparse Projected Residual Network (rsfMRI)}
SPRN processed the stability-selected rsfMRI features through successive hidden layers of widths 192, 96, 48, and 24. Each layer applied a linear transformation, batch normalization, GELU activation, and a dropout of 0.758; an identical dropout rate was enforced at the input. Learned linear projections supplied residual connections at the final three architectural transitions. Independent Gaussian noise characterized by a standard deviation of 0.2463 was injected into standardized imaging features exclusively during training. The final compact 24-dimensional representation was concatenated with covariates and forwarded to a linear output head.

\subsubsection{MAC-Net (tfMRI)}
MAC-Net represented each participant using 19 task conditions or contrasts across 94 regions. A shared linear projection mapped every contrast's 94 regional measurements into a 64-dimensional embedding, followed by SiLU activation. The resulting structured \(19\times64\) representation was flattened to 1,216 features and processed through sequential hidden layers of widths 256, 128, and 64, each equipped with batch normalization, SiLU, and a dropout of 0.7. Input dropout similarly utilized 0.7. Covariates were concatenated with the finalized imaging representation before passing through a linear regression head.

\subsection{Opportunistic Pretraining and Common-Cohort Adaptation}
The complete two-stage training and prediction fusion pipeline is depicted in Figure \ref{fig:two_stage_fusion}. For each target, configuration, and outer fold, Stage 1 trained each branch on common training participants and additional modality-eligible participants whose families were absent from the common validation and test partitions. Stage 2 initialized the model from its best Stage 1 checkpoint and used only common training participants for adaptation. The same common validation partition guided checkpoint selection in both stages. Training-derived numeric preprocessing and target transformations were retained during adaptation, and covariates were included through the branch-specific pathways described above.

Stage 1 used AdamW for up to 300 epochs, with early stopping after 30 epochs without improvement in validation $R^2$. Initial learning rates were $10^{-4}$ for IntelliGNN and $8.7\times10^{-4}$ for SPRN and MAC-Net. Weight decay was 0.005 for IntelliGNN and MAC-Net and 0.0071 for SPRN. Batch sizes were 64 for the two-stage IntelliGNN and 128 for SPRN and MAC-Net. The common-only IntelliGNN baseline used a batch size of 128. A plateau scheduler halved the learning rate after 10 epochs without improvement, subject to a minimum of $10^{-6}$.

During IntelliGNN adaptation, all parameters remained trainable. Graph-attention and node-embedding parameters used an initial learning rate of $5\times10^{-6}$, whereas fusion, pooling, and regression-head parameters used $5\times10^{-5}$. An $L_2$-SP penalty anchored the parameters to the selected Stage 1 checkpoint \cite{li2018explicit}:
\begin{equation}
\mathcal{L}_2(\theta)=\frac{1}{B}\sum_{i=1}^{B}\left(\widetilde{y}_i-f_\theta(x_i,c_i)\right)^2+0.01\sum_p\left\|\theta_p-\theta_p^{(1)}\right\|_2^2.
\end{equation}
Here, $\widetilde{y}_i$ is the transformed target, $x_i$ denotes imaging inputs, $c_i$ denotes covariates, and $\theta^{(1)}$ contains the selected Stage 1 parameters. The penalty included the regression-head parameters. A validation-$R^2$ plateau scheduler used patience 5, reduction factor 0.5, and minimum learning rate $10^{-6}$.

For SPRN and MAC-Net, only the terminal-head parameters received gradient updates during adaptation. AdamW was used with a learning rate of $5\times10^{-5}$ and cosine annealing over 100 epochs. Non-head parameters were frozen, but the models remained in training mode: dropout remained active, and batch-normalization running statistics were updated. SPRN also retained its training-time input noise. This procedure, therefore, combined head-parameter fine-tuning with adaptation of normalization statistics, rather than preserving a completely fixed feature mapping.

Stage 2 ran for at most 100 epochs with early-stopping patience 20. Its best checkpoint was retained only if its validation $R^2$ exceeded that of the selected Stage 1 checkpoint; otherwise, the Stage 1 checkpoint was restored. This fallback was determined exclusively by validation performance and did not guarantee improvement on the test partition. Random seeds were set to $42+1000k$ for outer fold $k$.

\subsection{Ablation Framework}
We evaluated pre-specified ablation variants in the leading configurations identified for each target to examine how alternative adaptation choices affected predictive performance. The variants were: (i) Stage 1 pretraining only, evaluated without Stage 2 adaptation; (ii) full-network versus head-only Stage 2 updating for the dense functional networks; and (iii) unregularized full-network and head-only adaptation for IntelliGNN. All variants used the same five-fold family-stratified partitions, target transformations, and validation-based checkpoint-selection criteria as the primary evaluation.


\subsection{Multimodal Prediction Fusion and Statistical Evaluation}

Prediction-level fusion was performed separately for each cognitive target, modality configuration, and outer fold. Predictions from the selected modality-specific checkpoints were generated for the common validation participants and inverse-transformed to the original cognitive-score scale. These predictions formed the inputs to a Ridge meta-regressor without additional standardization.

The regularization penalty, \(\alpha\), was selected using five-fold \texttt{GroupKFold} cross-validation within the common validation partition, keeping family members together within each internal split. The candidate set comprised 30 logarithmically spaced values between \(10^{-2}\) and \(10^{4}\), supplemented by \(\alpha=1\), yielding 31 candidates. Selection minimized mean validation MAE across the internal splits. The selected Ridge model was fitted to the common validation predictions and applied to the corresponding outer test predictions.

Because the common validation partition also guided neural checkpoint selection, this procedure constituted validation-set blending with internal family-grouped penalty selection. Outer test participants were excluded from neural training, checkpoint selection, and fusion-model fitting.

For statistical evaluation, outer test predictions were pooled across the five folds. The constituent comparator was selected post hoc as the branch with the lowest MAE across these pooled predictions. For participant \(i\), the paired absolute-error difference was defined as
\[
d_i =
|y_i-\widehat{y}_{i,\mathrm{base}}|
-
|y_i-\widehat{y}_{i,\mathrm{ensemble}}|,
\]
so positive values indicated lower ensemble error. The observed statistic was the mean of \(d_i\) across all pooled test participants, giving equal weight to each participant.

A family-aware sign-flipping test used 10,000 resamples with random seed 42. In each resample, an independently sampled sign was applied jointly to all differences belonging to the same family, and the participant-level mean was recalculated. The one-sided alternative was that the ensemble reduced MAE relative to the selected constituent. Reported \(p\)-values were nominal, without adjustment across target--configuration comparisons. The comparator selection was outcome-dependent and should be interpreted as a post hoc comparison of the evaluated predictors.

Performance summaries were calculated as unweighted means and population standard deviations across the five outer folds. Consequently, the pooled MAE difference used for statistical testing could differ slightly from the difference between reported fold-mean MAEs. Differences in \(R^2\), Pearson correlation, and MSE were reported descriptively.

\subsection{Predictive Evaluation}
Performance was evaluated using mean squared error (MSE), mean absolute error (MAE), Pearson correlation ($r$), and the coefficient of determination ($R^2$). Metrics were computed separately for each outer test fold using inverse-transformed predictions and the original observed cognitive scores. Validation checkpoint scores were calculated on the same original score scale. Observed outcomes were not reconstructed via a target-transform-and-inverse-transform cycle.

Results are reported as unweighted means and population standard deviations across five outer folds. Standard deviations describe between-fold variability; they are not confidence intervals or tests of differences between models. Within each configuration, predictors and training strategies were evaluated on the same saved test participants. Comparisons across configurations are descriptive because eligibility, training pools, and test participants differ.

The common-only versus two-stage comparison evaluates the complete training protocols, including opportunistic learning, adaptation, and validation-based checkpoint selection. It does not separately estimate the incremental effect of adaptation beyond pretraining. The IntelliGNN protocols also differed in batch size. Fusion comparisons assess the fitted ensemble relative to its constituent predictors and the covariate-only benchmarks within each configuration. The highest mean test $R^2$ among the five covariate-only model families is reported as a descriptive reference; it does not represent an additional test-independent model-selection procedure.

\subsection{Feature Attribution and Grouped Interpretation}
Selected modality-specific checkpoints were interpreted on the common outer test participants using Integrated Gradients (IG), signed Saliency, and Input $\times$ Gradient \cite{sundararajan2017axiomatic,simonyan2013deep,samek2021explaining}. Models were evaluated with dropout disabled and batch-normalization statistics fixed. Imaging inputs and covariates were attributed jointly. For IntelliGNN, graph topology and edge attributes remained fixed. Attributions explained each branch's transformed-target output before inverse transformation and Ridge fusion; they were not end-to-end explanations of the ensemble prediction.

IG used an all-zero reference in the preprocessed input space and a 50-step Gauss--Legendre approximation. Zero standardized continuous inputs represent training-derived reference values, whereas all-zero categorical indicators need not correspond to an observed participant profile. Signed Saliency was calculated with \texttt{abs=False}. For input $v$, reference $v'$, and feature $j$, the attribution quantities were
\begin{equation}
a_j^{\mathrm{IG}}(v)=(v_j-v'_j)\int_0^1\frac{\partial f(v'+\alpha(v-v'))}{\partial v_j}\,d\alpha,
\end{equation}
\begin{equation}
a_j^{\mathrm{SAL}}(v)=\frac{\partial f(v)}{\partial v_j},
\end{equation}
\begin{equation}
a_j^{\mathrm{IXG}}(v)=v_j\frac{\partial f(v)}{\partial v_j}.
\end{equation}
IG is a reference-relative path attribution, Saliency measures local sensitivity, and Input $\times$ Gradient scales that sensitivity by the input value.

Training-derived covariate and resting-feature preprocessing objects were reused. Structural scaling statistics were reconstructed from the designated training identifiers and target-specific feature mask. Structural Saliency entries corresponding to artificial padding or excluded feature groups were set to zero. Selected rsfMRI features were mapped to the complete feature layout, with unselected entries set to zero.

For method $q$, signed attributions were first averaged across the $n_k$ test participants in fold $k$ and then equally across folds:
\begin{equation}
\overline{a}_j^{(q)}=\frac{1}{5}\sum_{k=1}^{5}\left(\frac{1}{n_k}\sum_{i\in T_k}a_{ij}^{(q,k)}\right).
\end{equation}

Fold-specific attribution arrays were retained. For imaging inputs and covariates separately, the decoder averaged the stored signed attributions across folds, took the absolute value of each resulting mean, and min--max normalized these magnitudes within each method. The composite gradient-based attribution score was the arithmetic mean of the three normalized scores from IG, signed Saliency, and Input $\times$ Gradient:

\begin{equation}
\mathcal{N}(u)_j=\frac{u_j-\min_l u_l}{\max_l u_l-\min_l u_l+10^{-9}},
\end{equation}
\begin{equation}
I_j=\frac{1}{3}\sum_{q\in\{\mathrm{IG,SAL,IXG}\}}\mathcal{N}(|\overline{a}^{(q)}|)_j,
\end{equation}
\begin{equation}
d_j=\operatorname{sign}(\overline{a}_j^{(\mathrm{IG})}).
\end{equation}
Normalization was performed separately for imaging inputs and covariates within each model and target. Directions were obtained from IG alone. Taking magnitudes after signed averaging permits cancellation across participants and folds; the resulting score summarizes net attribution rather than mean absolute participant-level importance. The composite is a descriptive summary of related gradient-based methods, not an independent validation of their explanations.

Structural regional scores summed feature-level composite magnitudes within each node. Task regional scores were summed across conditions and contrasts. Resting-state regional scores summed incident-edge magnitudes and the corresponding regional variance magnitude, with each edge contributing to both endpoints. Regional directions were determined by the sign of the analogous sum of mean IG values. These aggregate scores can exceed 1 and are neither graph-centrality measures nor comparable effect sizes across models. Global edge rankings and connections restricted to the ten highest-ranked regions were reported separately.

For covariates, the sex and scanner indicator columns were summed to form composite scores, then min-max normalized. Group directions were calculated from the sum of the corresponding mean IG values. Categorical directions do not represent specific pairwise category contrasts, and grouped magnitude can depend on the number of indicator columns. Because imaging and covariates were normalized separately, their scores cannot be used to quantify relative contributions from imaging versus covariates.


Signed Ridge coefficients were averaged across outer folds to describe how the meta-regressor combined modality-specific predictions on the original cognitive-score scale. These unnormalized mean coefficients were reported separately from branch-level attributions. The decoder additionally generated min--max-normalized absolute mean coefficients, with the signs of the mean coefficients retained separately. For bimodal ensembles with unequal coefficient magnitudes, this normalization maps the smaller and larger magnitudes approximately to zero and one, respectively. A zero-normalized score, therefore, does not imply a zero regression coefficient. Neither the raw coefficients nor their normalized summaries represent proportions of explained variance.

\subsection{Implementation}
The attribution workflow used Python 3.10.20, PyTorch 2.5.1, Captum 0.9.0 \cite{captumsoftware2026}, and scikit-learn 1.7.1. Target and cohort identifiers were derived from the experiment directory structure. Data partitions, preprocessing objects, feature masks, graph definitions, model checkpoints, and fold-level attribution arrays were retained to support traceability and reproducibility in analysis.


\begin{table*}[t]
\centering
\small
\caption{Mean participant counts per outer fold. Common training, validation, and test participants belong to the configuration-specific intersection cohort. Architecture-specific Stage~1 pools include eligible common training participants and additional modality-eligible participants after exclusion of held-out families. Values are averages across five folds and are reported to one decimal place.}
\label{tab:opportunistic_counts}
\setlength{\tabcolsep}{10pt}
\resizebox{\textwidth}{!}{
\begin{tabular}{lllllll}
\toprule
Configuration & Common train & Common validation & Common test & IntelliGNN train & MAC-Net train & SPRN train \\
\midrule
sMRI+rsfMRI       & 3,426.2 & 849.0   & 1,068.8 & 7,578.0 & ---     & 3,629.6 \\
sMRI+tfMRI        & 4,381.6 & 1,094.4 & 1,369.0 & 7,054.0 & 4,805.4 & ---     \\
tfMRI+rsfMRI      & 2,793.6 & 691.2   & 871.2   & ---     & 5,616.6 & 3,925.0 \\
sMRI+tfMRI+rsfMRI & 2,698.6 & 674.2   & 843.2   & 7,953.8 & 5,657.0 & 3,966.2 \\
\bottomrule
\end{tabular}
}
\end{table*}

\section{Results}
\label{sec:results}

\subsection{Cohorts and Training Pools}
\label{sec:results_cohorts}
Seven partially overlapping imaging-eligibility cohorts were analyzed, ranging from 4,216 trimodal participants to 9,686 sMRI participants (Table \ref{tab:demographics_imputed}). Because eligibility, outcome distributions, and training pools differed across configurations, cross-configuration comparisons are descriptive.

Opportunistic pretraining increased the number of observations available to each modality-specific branch beyond the corresponding common training cohort (Table \ref{tab:opportunistic_counts}). In the trimodal configuration, the mean Stage 1 pools were approximately 2.95, 2.10, and 1.47 times the common training sample for IntelliGNN, MAC-Net, and SPRN, respectively. Complete imaging coverage was required for common-cohort validation and testing, but not for initial modality-specific training.

\subsection{Covariate, Unimodal, and Two-Stage Performance}
\label{sec:covariate_prediction}
Covariate-only models provided substantial predictive references across cohorts (Table \ref{tab:rerun_covariates}; Supplementary Tables S3a--S3c). The highest mean covariate-only \(R^2\) across model families ranged from 0.1417--0.1899 for \(G_f\), 0.2415--0.3020 for \(G_c\), and 0.2462--0.3016 for \(G_t\). These models included intracranial volume, applicable motion measures, and scanner variables; therefore, they quantify the added value of regional imaging features beyond this covariate set rather than beyond all MRI-derived information.

All nine unimodal imaging-plus-covariate models had a higher mean \(R^2\) than their corresponding covariate-only reference (Table \ref{tab:rerun_unimodal}; Supplementary Tables S4a--S4c). The increment varied by modality and target, ranging from 0.0121--0.0146 for sMRI to 0.0729 for rsfMRI predicting \(G_c\) and 0.0707 for tfMRI predicting \(G_t\). Differences in cohort composition and training-pool size preclude an intrinsic ranking of modality informativeness.

\begin{figure*}[htpb]
    \centering
    \includegraphics[width=\textwidth]{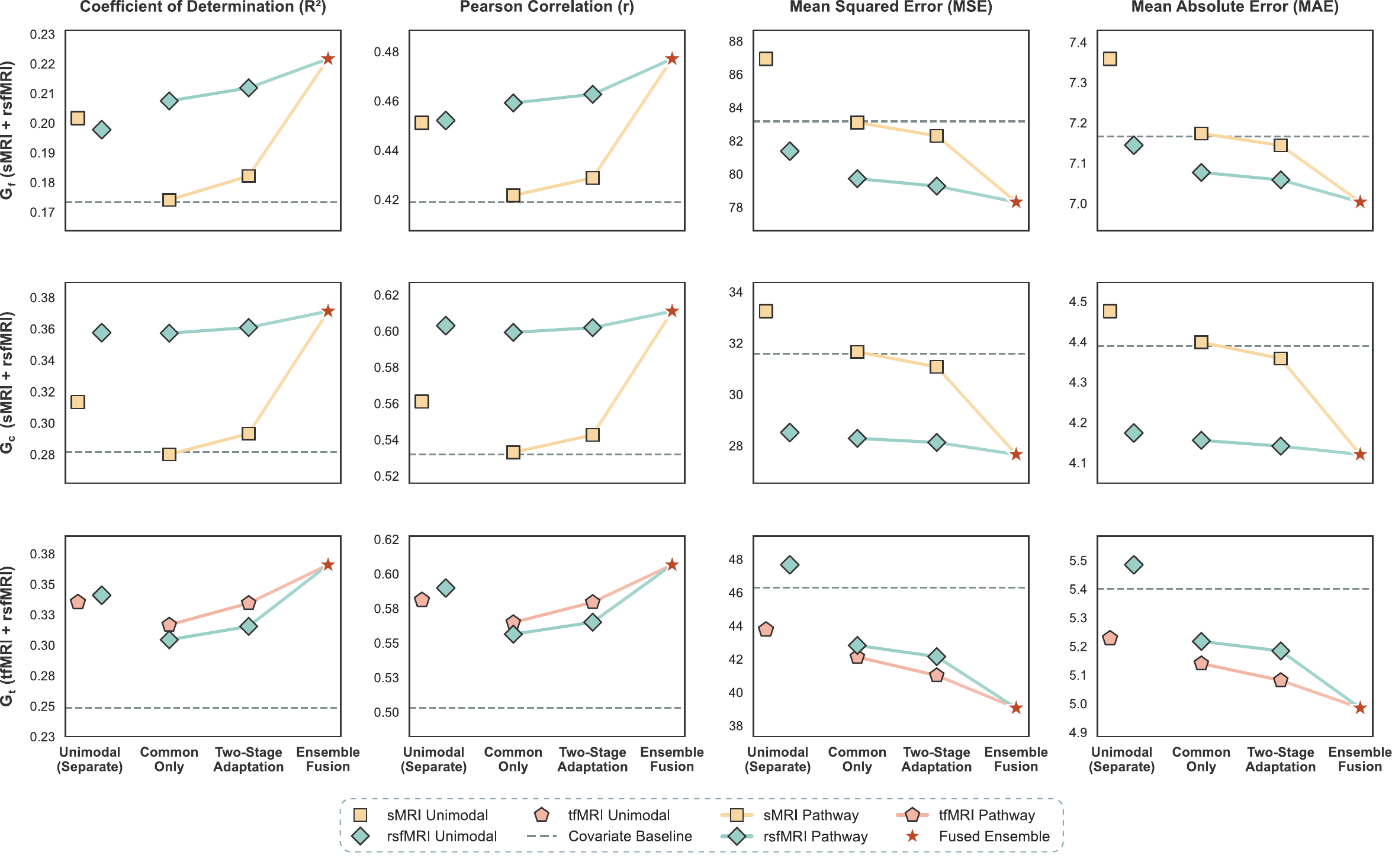}
    \caption{Mean predictive performance across training and fusion procedures for selected bimodal configurations: sMRI+rsfMRI for fluid and crystallized cognition and tfMRI+rsfMRI for total cognition. Panels show $R^2$, Pearson correlation, MSE, and MAE. The isolated unimodal points were obtained from the corresponding modality-eligible cohorts and therefore involve different participants. Common-only, two-stage, and ensemble points within a configuration share its test partitions. The two-stage points represent the complete opportunistic training and checkpoint-selection procedure, not an isolated adaptation effect. Dashed lines show the corresponding covariate benchmark. Differences are descriptive means across five folds; complete variability estimates are provided in the tables.}
    \label{fig:fusion_trajectories}
\end{figure*}

Within multimodal cohorts, the complete two-stage protocol had a higher mean \(R^2\) than common-cohort-only training in 24 of 27 architecture--target--configuration comparisons (Table \ref{tab:rerun_training}; Figure \ref{fig:fusion_trajectories}). Median differences were 0.0044 for \(G_f\) and 0.0089 for \(G_c\) and \(G_t\). MAC-Net produced several of the largest observed gains, including increases of 0.0233 for \(G_f\) in the tfMRI+rsfMRI configuration and 0.0259 for \(G_t\) in the trimodal configuration. Across its nine comparisons, changes ranged from -0.0016 to 0.0259. Three comparisons showed a small decrease or no difference in the reported precision. These results compare complete protocols, including opportunistic pretraining, common-cohort adaptation, and validation-based checkpoint selection; full four-metric results are reported in Supplementary Tables S4a--S4c.

\begin{table*}[t]
\centering
\caption{Covariate-only \(R^2\) across targets and eligibility cohorts.}
\label{tab:rerun_covariates}
\small
\setlength{\tabcolsep}{12pt}
\renewcommand{\arraystretch}{1.15}
\resizebox{\textwidth}{!}{
\begin{tabular}{lllllll}
\toprule
Target & Cohort & Ridge & Lasso & Elastic Net & Random forest & MLP \\
\midrule
\multirow{7}{*}{\(G_f\)} 
 & sMRI & \textbf{0.1899 \(\pm\) 0.0153} & 0.1888 \(\pm\) 0.0144 & 0.1854 \(\pm\) 0.0126 & 0.1686 \(\pm\) 0.0089 & 0.1857 \(\pm\) 0.0144 \\
 & rsfMRI & 0.1710 \(\pm\) 0.0191 & \textbf{0.1726 \(\pm\) 0.0185} & 0.1720 \(\pm\) 0.0182 & 0.1569 \(\pm\) 0.0165 & 0.1622 \(\pm\) 0.0207 \\
 & tfMRI & \textbf{0.1535 \(\pm\) 0.0053} & 0.1522 \(\pm\) 0.0055 & 0.1509 \(\pm\) 0.0043 & 0.1303 \(\pm\) 0.0138 & 0.1495 \(\pm\) 0.0066 \\
 & sMRI+rsfMRI & 0.1730 \(\pm\) 0.0191 & \textbf{0.1734 \(\pm\) 0.0188} & 0.1728 \(\pm\) 0.0187 & 0.1554 \(\pm\) 0.0174 & 0.1672 \(\pm\) 0.0275 \\
 & sMRI+tfMRI & 0.1448 \(\pm\) 0.0158 & 0.1479 \(\pm\) 0.0187 & \textbf{0.1485 \(\pm\) 0.0182} & 0.1268 \(\pm\) 0.0229 & 0.1411 \(\pm\) 0.0177 \\
 & tfMRI+rsfMRI & 0.1411 \(\pm\) 0.0265 & \textbf{0.1417 \(\pm\) 0.0261} & 0.1403 \(\pm\) 0.0251 & 0.1175 \(\pm\) 0.0226 & 0.1322 \(\pm\) 0.0287 \\
 & sMRI+tfMRI+rsfMRI & 0.1375 \(\pm\) 0.0113 & 0.1379 \(\pm\) 0.0127 & 0.1387 \(\pm\) 0.0117 & 0.1171 \(\pm\) 0.0053 & \textbf{0.1434 \(\pm\) 0.0056} \\
\midrule
\multirow{7}{*}{\(G_c\)} 
 & sMRI & \textbf{0.3020 \(\pm\) 0.0146} & 0.3015 \(\pm\) 0.0162 & 0.2983 \(\pm\) 0.0155 & 0.2819 \(\pm\) 0.0171 & 0.3012 \(\pm\) 0.0128 \\
 & rsfMRI & 0.2826 \(\pm\) 0.0220 & \textbf{0.2847 \(\pm\) 0.0220} & 0.2823 \(\pm\) 0.0218 & 0.2577 \(\pm\) 0.0224 & 0.2794 \(\pm\) 0.0241 \\
 & tfMRI & 0.2619 \(\pm\) 0.0116 & 0.2625 \(\pm\) 0.0109 & 0.2590 \(\pm\) 0.0089 & 0.2433 \(\pm\) 0.0048 & \textbf{0.2628 \(\pm\) 0.0133} \\
 & sMRI+rsfMRI & \textbf{0.2816 \(\pm\) 0.0153} & 0.2812 \(\pm\) 0.0150 & 0.2774 \(\pm\) 0.0136 & 0.2673 \(\pm\) 0.0161 & 0.2777 \(\pm\) 0.0194 \\
 & sMRI+tfMRI & 0.2498 \(\pm\) 0.0090 & \textbf{0.2512 \(\pm\) 0.0111} & 0.2473 \(\pm\) 0.0112 & 0.2393 \(\pm\) 0.0216 & 0.2506 \(\pm\) 0.0123 \\
 & tfMRI+rsfMRI & 0.2406 \(\pm\) 0.0141 & \textbf{0.2422 \(\pm\) 0.0116} & 0.2419 \(\pm\) 0.0151 & 0.2283 \(\pm\) 0.0208 & 0.2385 \(\pm\) 0.0156 \\
 & sMRI+tfMRI+rsfMRI & 0.2414 \(\pm\) 0.0097 & \textbf{0.2415 \(\pm\) 0.0115} & 0.2401 \(\pm\) 0.0090 & 0.2158 \(\pm\) 0.0181 & 0.2358 \(\pm\) 0.0179 \\
\midrule
\multirow{7}{*}{\(G_t\)} 
 & sMRI & \textbf{0.3016 \(\pm\) 0.0110} & 0.3009 \(\pm\) 0.0113 & 0.2949 \(\pm\) 0.0101 & 0.2784 \(\pm\) 0.0128 & 0.2986 \(\pm\) 0.0130 \\
 & rsfMRI & 0.2839 \(\pm\) 0.0209 & 0.2859 \(\pm\) 0.0217 & \textbf{0.2864 \(\pm\) 0.0223} & 0.2646 \(\pm\) 0.0232 & 0.2814 \(\pm\) 0.0241 \\
 & tfMRI & 0.2620 \(\pm\) 0.0117 & \textbf{0.2646 \(\pm\) 0.0108} & 0.2615 \(\pm\) 0.0133 & 0.2374 \(\pm\) 0.0171 & 0.2603 \(\pm\) 0.0103 \\
 & sMRI+rsfMRI & \textbf{0.2846 \(\pm\) 0.0237} & 0.2823 \(\pm\) 0.0255 & 0.2816 \(\pm\) 0.0243 & 0.2670 \(\pm\) 0.0215 & 0.2778 \(\pm\) 0.0270 \\
 & sMRI+tfMRI & \textbf{0.2525 \(\pm\) 0.0194} & 0.2524 \(\pm\) 0.0183 & 0.2508 \(\pm\) 0.0196 & 0.2325 \(\pm\) 0.0292 & 0.2478 \(\pm\) 0.0180 \\
 & tfMRI+rsfMRI & \textbf{0.2485 \(\pm\) 0.0216} & 0.2484 \(\pm\) 0.0195 & 0.2474 \(\pm\) 0.0193 & 0.2277 \(\pm\) 0.0188 & 0.2442 \(\pm\) 0.0305 \\
 & sMRI+tfMRI+rsfMRI & 0.2459 \(\pm\) 0.0118 & \textbf{0.2462 \(\pm\) 0.0103} & 0.2452 \(\pm\) 0.0080 & 0.2218 \(\pm\) 0.0091 & 0.2449 \(\pm\) 0.0080 \\
\bottomrule
\end{tabular}%
}
\end{table*}

\begin{table*}[t]
\centering
\caption{Unimodal brain-plus-covariate performance. Bold values indicate the best-performing architecture for a given metric within each cognitive target (highest for \(R^2\), Pearson r, and \(\Delta R^2\); lowest for MSE and MAE).}
\label{tab:rerun_unimodal}
\small
\setlength{\tabcolsep}{14pt}
\renewcommand{\arraystretch}{1.12}
\resizebox{\textwidth}{!}{%
\begin{tabular}{lllllll}
\toprule
Target & Branch & \(R^2\) & Pearson r & MSE & MAE & \(\Delta R^2\) (cov.) \\
\midrule
\multirow{3}{*}{\(G_f\)} 
 & IntelliGNN & \textbf{0.2023 \(\pm\) 0.0153} & 0.4530 \(\pm\) 0.0134 & 86.8566 \(\pm\) 2.3671 & 7.3535 \(\pm\) 0.0929 & +0.0124 \\
 & SPRN & 0.1978 \(\pm\) 0.0295 & 0.4521 \(\pm\) 0.0297 & 81.3778 \(\pm\) 5.2820 & 7.1445 \(\pm\) 0.2368 & +0.0252 \\
 & MAC-Net & 0.2007 \(\pm\) 0.0208 & \textbf{0.4535 \(\pm\) 0.0186} & \textbf{73.2790 \(\pm\) 4.1989} & \textbf{6.7683 \(\pm\) 0.2165} & \textbf{+0.0472} \\
\midrule
\multirow{3}{*}{\(G_c\)} 
 & IntelliGNN & 0.3166 \(\pm\) 0.0176 & 0.5635 \(\pm\) 0.0162 & 33.1018 \(\pm\) 0.8867 & 4.4649 \(\pm\) 0.0607 & +0.0146 \\
 & SPRN & \textbf{0.3576 \(\pm\) 0.0143} & \textbf{0.6030 \(\pm\) 0.0159} & \textbf{28.5155 \(\pm\) 1.9414} & \textbf{4.1737 \(\pm\) 0.1500} & \textbf{+0.0729} \\
 & MAC-Net & 0.3114 \(\pm\) 0.0113 & 0.5591 \(\pm\) 0.0103 & 29.5441 \(\pm\) 1.3299 & 4.2313 \(\pm\) 0.0841 & +0.0486 \\
\midrule
\multirow{3}{*}{\(G_t\)} 
 & IntelliGNN & 0.3137 \(\pm\) 0.0130 & 0.5622 \(\pm\) 0.0102 & 54.8532 \(\pm\) 1.3464 & 5.8370 \(\pm\) 0.0694 & +0.0121 \\
 & SPRN & \textbf{0.3410 \(\pm\) 0.0270} & \textbf{0.5898 \(\pm\) 0.0260} & 47.6685 \(\pm\) 3.1669 & 5.4846 \(\pm\) 0.1809 & +0.0546 \\
 & MAC-Net & 0.3353 \(\pm\) 0.0242 & 0.5811 \(\pm\) 0.0190 & \textbf{43.7696 \(\pm\) 1.8478} & \textbf{5.2277 \(\pm\) 0.1354} & \textbf{+0.0707} \\
\bottomrule
\end{tabular}%
}
\end{table*}

\begin{table*}[t]
\centering
\caption{Mean test \(R^2\) for common-only neural training and the complete two-stage strategy within each multimodal cohort. Each comparison uses the same configuration-specific test participants. The two-stage results include opportunistic pretraining, common-cohort adaptation, and validation-based checkpoint selection; the comparison does not isolate adaptation alone.}
\label{tab:rerun_training}
\small
\setlength{\tabcolsep}{28pt}
\renewcommand{\arraystretch}{1.12}
\resizebox{\textwidth}{!}{%
\begin{tabular}{lllll}
\toprule
Cohort & Base model & \(G_f\) & \(G_c\) & \(G_t\) \\
\midrule
\multirow{2}{*}{sMRI+rsfMRI} 
 & IntelliGNN & 0.1741 \(\rightarrow\) 0.1822 & 0.2801 \(\rightarrow\) 0.2934 & 0.2861 \(\rightarrow\) 0.2950 \\
 & SPRN & 0.2075 \(\rightarrow\) 0.2119 & 0.3573 \(\rightarrow\) 0.3609 & 0.3513 \(\rightarrow\) 0.3513 \\
\midrule
\multirow{2}{*}{sMRI+tfMRI} 
 & IntelliGNN & 0.1517 \(\rightarrow\) 0.1551 & 0.2596 \(\rightarrow\) 0.2620 & 0.2565 \(\rightarrow\) 0.2695 \\
 & MAC-Net & 0.2052 \(\rightarrow\) 0.2036 & 0.3065 \(\rightarrow\) 0.3100 & 0.3331 \(\rightarrow\) 0.3358 \\
\midrule
\multirow{2}{*}{tfMRI+rsfMRI} 
 & SPRN & 0.1664 \(\rightarrow\) 0.1747 & 0.3260 \(\rightarrow\) 0.3242 & 0.3045 \(\rightarrow\) 0.3155 \\
 & MAC-Net & 0.1744 \(\rightarrow\) 0.1977 & 0.2892 \(\rightarrow\) 0.3086 & 0.3167 \(\rightarrow\) 0.3345 \\
\midrule
\multirow{3}{*}{sMRI+tfMRI+rsfMRI} 
 & IntelliGNN & 0.1369 \(\rightarrow\) 0.1381 & 0.2412 \(\rightarrow\) 0.2520 & 0.2498 \(\rightarrow\) 0.2505 \\
 & SPRN & 0.1552 \(\rightarrow\) 0.1588 & 0.3132 \(\rightarrow\) 0.3221 & 0.2932 \(\rightarrow\) 0.2997 \\
 & MAC-Net & 0.1839 \(\rightarrow\) 0.2051 & 0.2799 \(\rightarrow\) 0.3009 & 0.3094 \(\rightarrow\) 0.3353 \\
\bottomrule
\end{tabular}%
}
\end{table*}

\subsection{Multimodal Fusion: Consistent Within-Cohort Gains and Target-Specific Trade-offs}
\label{sec:multimodal_fusion}
All 12 ensembles had higher mean \(R^2\) and Pearson correlation and lower mean MSE and MAE than their strongest constituent predictors. Family-aware sign-flipping tests provided nominal evidence of MAE reductions for tfMRI+rsfMRI and trimodal fusion across all three targets (\(p \leq 0.0108\)). For sMRI+rsfMRI, reductions were nominally significant for \(G_f\) (\(p=0.0012\)) and \(G_t\) (\(p=0.0001\)), but not for \(G_c\) (\(p=0.0626\)). The sMRI+tfMRI comparisons did not reach nominal significance for \(G_f\) (\(p=0.0802\)), \(G_c\) (\(p=0.0597\)), or \(G_t\) (\(p=0.1876\)). These tests assess absolute-error reduction within each configuration; they do not establish differences between configurations or identify the mechanism underlying the fusion gains.


Across independent configurations, the highest-performing observed configuration depended on the target and evaluation metric. For \(G_f\), tfMRI+rsfMRI and sMRI+rsfMRI performed similarly in terms of explained variance (\(R^2\) = 0.2222 vs. 0.2216). sMRI+rsfMRI achieved a marginally higher correlation (0.4773 vs. 0.4743), whereas tfMRI+rsfMRI generated substantially lower MSE (68.6384 vs. 78.3221) and MAE (6.5746 vs. 7.0037). The difference of 0.0006 in mean \(R^2\) was small relative to the observed between-fold variability and does not establish a performance difference or equivalence between the two configurations.

For \(G_c\), sMRI+rsfMRI achieved the highest \(R^2\) (0.3706) and correlation (0.6108), yet it did not absolutely minimize absolute error. tfMRI+rsfMRI recorded the lowest MAE (3.9901), while the fully trimodal sMRI+tfMRI+rsfMRI ensemble achieved the lowest overall MSE (25.7145). The trimodal model's slightly lower MSE compared to tfMRI+rsfMRI coexisted with a lower \(R^2\), lower correlation, and a marginally higher MAE. For \(G_t\), tfMRI+rsfMRI performed the best across all four metrics, achieving the highest \(R^2\) (0.3665) and correlation (0.6069), alongside the lowest MSE (39.0299) and MAE (4.9836). These observed mean performances motivated selecting tfMRI+rsfMRI for \(G_f\) and \(G_t\), and sMRI+rsfMRI for \(G_c\), for detailed explanatory analysis.


No trimodal ensemble achieved the highest observed mean \(R^2\) for any target. Nevertheless, fusion improved the reported mean metrics over its strongest constituent within each configuration. Because configurations differed in eligible participants, outcome distributions, and training pools, cross-configuration rankings cannot isolate the effect of adding or removing a modality.

\begin{table*}[t]
\centering
\caption{Multimodal ensemble performance using validation-set Ridge blending with internal family-grouped penalty selection.}
\label{tab:rerun_ensemble}
\small
\setlength{\tabcolsep}{18pt}
\renewcommand{\arraystretch}{1.12}
\resizebox{\textwidth}{!}{%
\begin{tabular}{llllll}
\toprule
Target & Cohort & \(R^2\) & Pearson r & MSE & MAE \\
\midrule
\multirow{4}{*}{\(G_f\)} 
 & sMRI+rsfMRI & 0.2216 \(\pm\) 0.0146 & \textbf{0.4773 \(\pm\) 0.0185} & 78.3221 \(\pm\) 2.6582 & 7.0037 \(\pm\) 0.1642 \\
 & sMRI+tfMRI & 0.2095 \(\pm\) 0.0175 & 0.4581 \(\pm\) 0.0195 & 71.7956 \(\pm\) 3.5191 & 6.7300 \(\pm\) 0.1362 \\
 & tfMRI+rsfMRI & \textbf{0.2222 \(\pm\) 0.0202} & 0.4743 \(\pm\) 0.0211 & \textbf{68.6384 \(\pm\) 4.2938} & \textbf{6.5746 \(\pm\) 0.2131} \\
 & sMRI+tfMRI+rsfMRI & 0.2211 \(\pm\) 0.0174 & 0.4719 \(\pm\) 0.0169 & 68.7092 \(\pm\) 3.1316 & 6.5962 \(\pm\) 0.1703 \\
\midrule
\multirow{4}{*}{\(G_c\)} 
 & sMRI+rsfMRI & \textbf{0.3706 \(\pm\) 0.0162} & \textbf{0.6108 \(\pm\) 0.0140} & 27.6907 \(\pm\) 1.5092 & 4.1229 \(\pm\) 0.1054 \\
 & sMRI+tfMRI & 0.3167 \(\pm\) 0.0179 & 0.5632 \(\pm\) 0.0163 & 29.2300 \(\pm\) 1.5637 & 4.2356 \(\pm\) 0.0979 \\
 & tfMRI+rsfMRI & 0.3575 \(\pm\) 0.0158 & 0.5994 \(\pm\) 0.0139 & 25.8052 \(\pm\) 0.7025 & \textbf{3.9901 \(\pm\) 0.0561} \\
 & sMRI+tfMRI+rsfMRI & 0.3548 \(\pm\) 0.0098 & 0.5976 \(\pm\) 0.0075 & \textbf{25.7145 \(\pm\) 0.7188} & 3.9990 \(\pm\) 0.0607 \\
\midrule
\multirow{4}{*}{\(G_t\)} 
 & sMRI+rsfMRI & 0.3648 \(\pm\) 0.0255 & 0.6068 \(\pm\) 0.0222 & 45.4731 \(\pm\) 1.7370 & 5.3480 \(\pm\) 0.1033 \\
 & sMRI+tfMRI & 0.3409 \(\pm\) 0.0175 & 0.5844 \(\pm\) 0.0153 & 42.9730 \(\pm\) 1.7159 & 5.1978 \(\pm\) 0.0747 \\
 & tfMRI+rsfMRI & \textbf{0.3665 \(\pm\) 0.0131} & \textbf{0.6069 \(\pm\) 0.0118} & \textbf{39.0299 \(\pm\) 2.1194} & \textbf{4.9836 \(\pm\) 0.1417} \\
 & sMRI+tfMRI+rsfMRI & 0.3615 \(\pm\) 0.0142 & 0.6032 \(\pm\) 0.0112 & 39.0602 \(\pm\) 1.6983 & 4.9846 \(\pm\) 0.0944 \\
\bottomrule
\end{tabular}%
}
\end{table*}

\begin{table*}[t]
\centering
\caption{Within-cohort fusion gains and family-aware statistical significance testing. \(\Delta\) values represent the ensemble mean minus the strongest constituent base mean. Positive values for \(R^2\) and Pearson \(r\) indicate predictive improvement; negative values for MSE and MAE indicate error reduction. The \(p\)-values are from one-sided family-aware sign-flipping tests of paired absolute-error differences and are interpreted as nominal evidence across the 12 comparisons.}
\label{tab:rerun_gains}
\small
\setlength{\tabcolsep}{18pt}
\renewcommand{\arraystretch}{1.12}
\resizebox{\textwidth}{!}{
\begin{tabular}{llllllll}
\toprule
Target & Cohort & Best base & \(\Delta R^2\) & \(\Delta r\) & \(\Delta\) MSE & \(\Delta\) MAE & \(p\)-value \\
\midrule
\multirow{4}{*}{\(G_f\)} 
 & sMRI+rsfMRI & SPRN & +0.0097 & +0.0146 & -0.9566 & -0.0547 & 0.0012 \\
 & sMRI+tfMRI & MAC-Net & +0.0059 & +0.0040 & -0.5293 & -0.0138 & 0.0802 \\
 & tfMRI+rsfMRI & MAC-Net & +0.0245 & +0.0252 & -2.1959 & -0.0914 & <0.0001 \\
 & sMRI+tfMRI+rsfMRI & MAC-Net & +0.0160 & +0.0159 & -1.4104 & -0.0526 & 0.0108 \\
\midrule
\multirow{4}{*}{\(G_c\)} 
 & sMRI+rsfMRI & SPRN & +0.0097 & +0.0089 & -0.4382 & -0.0182 & 0.0626 \\
 & sMRI+tfMRI & MAC-Net & +0.0067 & +0.0042 & -0.2874 & -0.0116 & 0.0597 \\
 & tfMRI+rsfMRI & SPRN & +0.0333 & +0.0241 & -1.3479 & -0.0579 & 0.0002 \\
 & sMRI+tfMRI+rsfMRI & SPRN & +0.0327 & +0.0253 & -1.3043 & -0.0659 & 0.0015 \\
\midrule
\multirow{4}{*}{\(G_t\)} 
 & sMRI+rsfMRI & SPRN & +0.0135 & +0.0123 & -0.9743 & -0.0551 & 0.0001 \\
 & sMRI+tfMRI & MAC-Net & +0.0051 & +0.0029 & -0.3300 & -0.0067 & 0.1876 \\
 & tfMRI+rsfMRI & MAC-Net & +0.0320 & +0.0274 & -1.9888 & -0.0976 & <0.0001 \\
 & sMRI+tfMRI+rsfMRI & MAC-Net & +0.0262 & +0.0226 & -1.6211 & -0.0679 & 0.0004 \\
\bottomrule
\end{tabular}%
}
\end{table*}

\begin{table}[t]
\centering\small
\caption{Mean signed RidgeCV coefficients for prediction-level fusion. Coefficients act on score-scale predictions without additional standardization.}
\label{tab:modality_coefficients}
\setlength{\tabcolsep}{10pt}
\renewcommand{\arraystretch}{1.13}
\resizebox{\linewidth}{!}{
\begin{tabular}{lllll}
\toprule
Target & Cohort & sMRI & rsfMRI & tfMRI \\
\midrule
\multirow{4}{*}{\(G_f\)} 
 & sMRI+rsfMRI & 0.438 & 0.644 & --- \\
 & sMRI+tfMRI & 0.240 & --- & 0.764 \\
 & tfMRI+rsfMRI & --- & 0.397 & 0.683 \\
 & sMRI+tfMRI+rsfMRI & 0.085 & 0.402 & 0.634 \\
\midrule
\multirow{4}{*}{\(G_c\)} 
 & sMRI+rsfMRI & 0.374 & 0.737 & --- \\
 & sMRI+tfMRI & 0.271 & --- & 0.758 \\
 & tfMRI+rsfMRI & --- & 0.618 & 0.445 \\
 & sMRI+tfMRI+rsfMRI & -0.134 & 0.623 & 0.633 \\
\midrule
\multirow{4}{*}{\(G_t\)} 
 & sMRI+rsfMRI & 0.375 & 0.699 & --- \\
 & sMRI+tfMRI & 0.206 & --- & 0.820 \\
 & tfMRI+rsfMRI & --- & 0.458 & 0.607 \\
 & sMRI+tfMRI+rsfMRI & -0.043 & 0.470 & 0.713 \\
\bottomrule
\end{tabular}
}
\end{table}

\begin{figure*}[htpb]
    \centering
    \includegraphics[width=\textwidth]{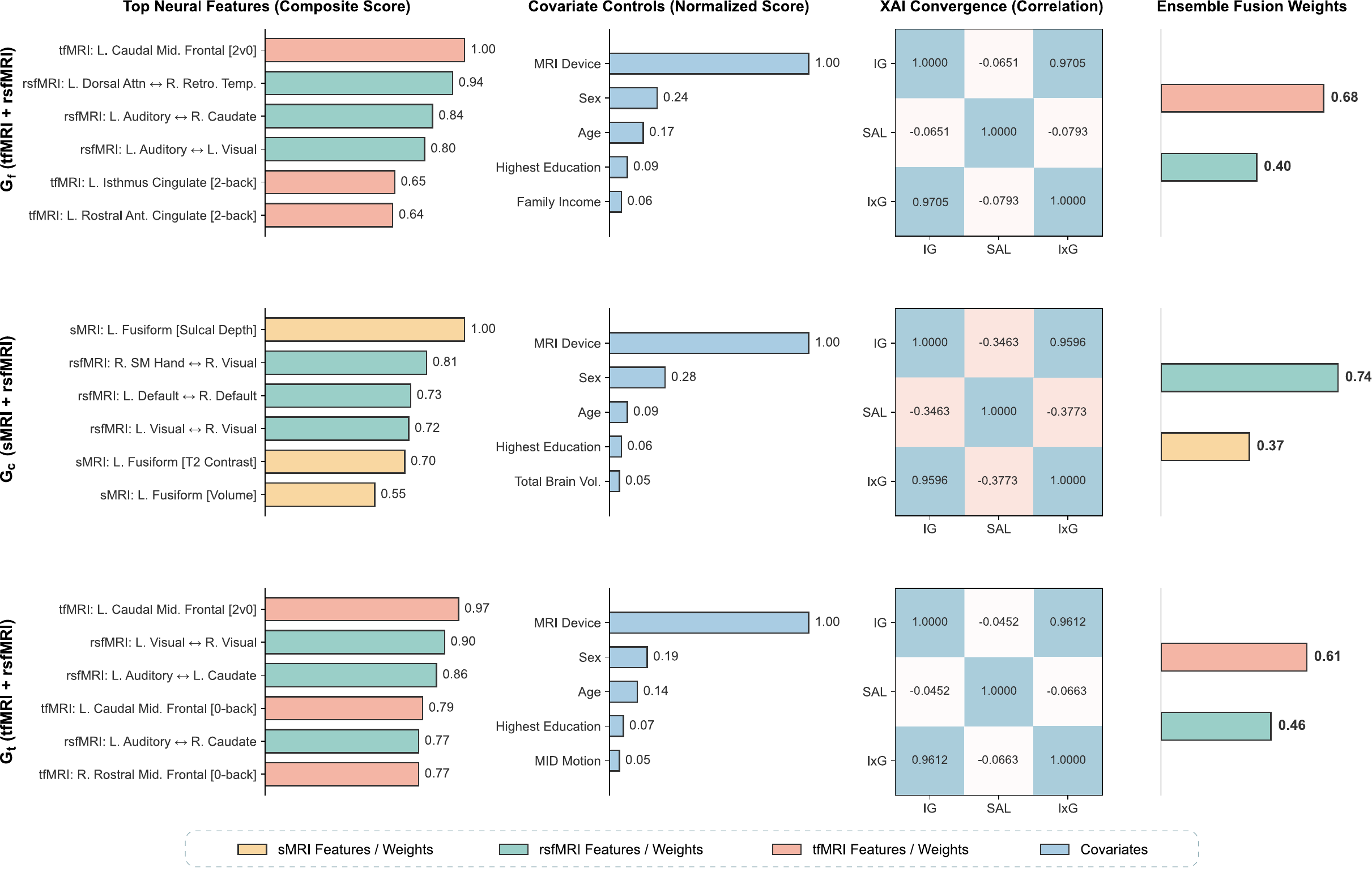}
    \caption{Explainability and late-fusion summaries for the selected multimodal pipelines across fluid ($G_f$), crystallized ($G_c$), and total ($G_t$) cognition. Column 1 displays the top six neuroanatomical and connectomic features ranked using cross-fold composite gradient-based attribution scores. Feature scores were normalized separately within each constituent branch from Integrated Gradients (IG), signed Saliency, and Input $\times$ Gradient; their displayed magnitudes do not quantify relative importance across modalities. Column 2 displays grouped covariate scores, obtained by summing categorical indicator scores and min--max normalizing the grouped values; scanner identity has the highest score in the displayed configurations. Column 3 presents agreement among method-specific attribution magnitudes, with high correlation ($r > 0.95$) between IG and Input $\times$ Gradient, and distinct patterns for Saliency. This agreement describes method-specific consistency rather than independent validation of the extracted features. Column 4 displays the cross-fold mean signed RidgeCV coefficients acting on score-scale modality-specific predictions, rounded to two decimal places. These coefficients are not normalized proportions and do not represent percentages of explained variance.}
    \label{fig:xai_findings}
\end{figure*}

\subsection{Fusion Weights and Attribution Summaries}
Mean Ridge coefficients varied by configuration (Table \ref{tab:modality_coefficients}). In bimodal ensembles, sMRI coefficients were positive but smaller than the functional coefficient. Within tfMRI+rsfMRI, tfMRI had the larger coefficient for \(G_f\) and \(G_t\), whereas rsfMRI had the larger coefficient for \(G_c\). In trimodal models, the mean sMRI coefficient was small and positive for \(G_f\) (0.085) and negative for \(G_c\) (-0.134) and \(G_t\) (-0.043). These conditional regression weights describe how correlated score-scale predictions were combined; they are not proportions of modality-specific biological variance.

Figure \ref{fig:xai_findings} summarises attribution patterns in the selected leading configurations, and complete rankings are reported in Supplementary Tables S5--S8. For \(G_f\), the tfMRI+rsfMRI model assigned the highest task-feature attribution to the left caudal middle frontal 2-back versus 0-back contrast. For \(G_c\), left fusiform structural measures and distributed rsfMRI features across default-mode, sensory, and attention-related systems ranked highly. For \(G_t\), the left caudal middle frontal 2-back versus 0-back contrast again ranked highly, together with visual and auditory--caudate resting-state connections.

Across configurations, bilateral fusiform regions recurred in structural regional summaries, while tfMRI rankings for \(G_f\) and \(G_t\) were dominated by N-back measures; \(G_c\) rankings additionally included MID loss-anticipation contrasts. Resting-state summaries repeatedly included default, sensory, and attention-network labels across configurations. Scanner identity had the highest normalized grouped composite covariate score in all 27 summaries. These attribution results describe features emphasized by the fitted models and should not be interpreted as causal evidence for neurobiological mechanisms.

Integrated Gradients and Input \(\times\) Gradient attribution magnitudes were strongly correlated in the selected configurations (\(r\geq0.9596\)), whereas Saliency differed. This finding indicates that attribution rankings depend on the selected formulation. Agreement between the two related gradient-based methods is a sensitivity result and does not independently validate biological relevance or the stability of explanations across model retraining.

\subsection{Ablation Results}

In the selected configurations, Stage-1-only checkpoints retained performance close to that of the complete two-stage protocol (Table \ref{tab:ablation_results}), indicating that expanded Stage 1 training accounted for much of the observed difference relative to common-cohort-only training. For example, SPRN for \(G_c\) had the same reported mean \(R^2\) with Stage 1 only and the complete protocol (0.3609), whereas MAC-Net for \(G_t\) had mean \(R^2\) values of 0.3348 and 0.3345, respectively.

Full-network Stage 2 updating did not consistently outperform head-only updating in the dense networks. For MAC-Net predicting \(G_f\), full-network updating had a higher mean \(R^2\) than head-only updating (0.2001 versus 0.1977), whereas the ordering was reversed for \(G_t\) (0.3322 versus 0.3345). SPRN favored head-only updating for \(G_f\) and \(G_t\), while its \(G_c\) results were effectively unchanged at the reported precision. For IntelliGNN, regularized full-network adaptation had the highest observed mean \(R^2\) among the tested variants (0.2934). These ablations suggest architecture-specific transfer behavior within the evaluated configurations; they do not establish a general mechanism of representation drift or catastrophic forgetting.

\begin{table*}[t]
\centering
\caption{Ablation analysis across optimal target configurations under 5-fold cross-validation. Metrics reflect unweighted means \(\pm\) standard deviations across outer test folds. Bold values indicate the best-performing metric within each ablation group for its modality.}
\label{tab:ablation_results}
\small
\setlength{\tabcolsep}{10pt}
\renewcommand{\arraystretch}{1.12}
\resizebox{\textwidth}{!}{%
\begin{tabular}{llllcccc}
\toprule
Target & Modality & Model & Adaptation Regime & \(R^2\) & Pearson \(r\) & MSE & MAE \\
\midrule
\multirow{6}{*}{\(G_f\)} 
 & \multirow{6}{*}{tfMRI+rsfMRI} 
   & \multirow{3}{*}{MAC-Net} 
     & S1 Only & 0.2000 \(\pm\) 0.0177 & \textbf{0.4509 \(\pm\) 0.0191} & 70.6238 \(\pm\) 4.7386 & \textbf{6.6545 \(\pm\) 0.2285} \\
 & & & S2 Unfrozen & \textbf{0.2001 \(\pm\) 0.0184} & 0.4506 \(\pm\) 0.0198 & \textbf{70.6216 \(\pm\) 4.8065} & 6.6581 \(\pm\) 0.2366 \\
 & & & S2 Frozen (Prop.) & 0.1977 \(\pm\) 0.0177 & 0.4491 \(\pm\) 0.0187 & 70.8343 \(\pm\) 4.8362 & 6.6660 \(\pm\) 0.2357 \\
\cmidrule{3-8}
 & & \multirow{3}{*}{SPRN} 
     & S1 Only & 0.1713 \(\pm\) 0.0376 & 0.4297 \(\pm\) 0.0286 & 73.0565 \(\pm\) 4.2562 & 6.8135 \(\pm\) 0.2146 \\
 & & & S2 Unfrozen & 0.1724 \(\pm\) 0.0356 & 0.4295 \(\pm\) 0.0286 & 72.9624 \(\pm\) 4.1268 & 6.8093 \(\pm\) 0.2070 \\
 & & & S2 Frozen (Prop.) & \textbf{0.1747 \(\pm\) 0.0346} & \textbf{0.4305 \(\pm\) 0.0281} & \textbf{72.7745 \(\pm\) 4.2558} & \textbf{6.7952 \(\pm\) 0.2165} \\
\midrule
\multirow{7}{*}{\(G_c\)} 
 & \multirow{7}{*}{sMRI+rsfMRI} 
   & \multirow{4}{*}{IntelliGNN} 
     & S1 Only & 0.2920 \(\pm\) 0.0155 & 0.5416 \(\pm\) 0.0140 & 31.1446 \(\pm\) 1.5130 & 4.3608 \(\pm\) 0.1059 \\
 & & & S2 No \(L_2\)-SP & 0.2920 \(\pm\) 0.0155 & 0.5416 \(\pm\) 0.0140 & 31.1446 \(\pm\) 1.5130 & 4.3608 \(\pm\) 0.1059 \\
 & & & S2 Frozen & 0.2927 \(\pm\) 0.0147 & 0.5426 \(\pm\) 0.0129 & 31.1112 \(\pm\) 1.4671 & \textbf{4.3566 \(\pm\) 0.1020} \\
 & & & S2 \(L_2\)-SP (Prop.) & \textbf{0.2934 \(\pm\) 0.0140} & \textbf{0.5427 \(\pm\) 0.0128} & \textbf{31.0790 \(\pm\) 1.4264} & 4.3582 \(\pm\) 0.1035 \\
\cmidrule{3-8}
 & & \multirow{3}{*}{SPRN} 
     & S1 Only & \textbf{0.3609 \(\pm\) 0.0178} & \textbf{0.6019 \(\pm\) 0.0150} & \textbf{28.1288 \(\pm\) 1.7054} & \textbf{4.1411 \(\pm\) 0.1138} \\
 & & & S2 Unfrozen & \textbf{0.3609 \(\pm\) 0.0178} & \textbf{0.6019 \(\pm\) 0.0150} & \textbf{28.1288 \(\pm\) 1.7054} & \textbf{4.1411 \(\pm\) 0.1138} \\
 & & & S2 Frozen (Prop.) & \textbf{0.3609 \(\pm\) 0.0178} & \textbf{0.6019 \(\pm\) 0.0150} & 28.1289 \(\pm\) 1.7053 & \textbf{4.1411 \(\pm\) 0.1138} \\
\midrule
\multirow{6}{*}{\(G_t\)} 
 & \multirow{6}{*}{tfMRI+rsfMRI} 
   & \multirow{3}{*}{MAC-Net} 
     & S1 Only & \textbf{0.3348 \(\pm\) 0.0268} & \textbf{0.5796 \(\pm\) 0.0221} & \textbf{41.0134 \(\pm\) 3.0078} & \textbf{5.0802 \(\pm\) 0.1760} \\
 & & & S2 Unfrozen & 0.3322 \(\pm\) 0.0253 & 0.5774 \(\pm\) 0.0207 & 41.1638 \(\pm\) 2.9016 & 5.0855 \(\pm\) 0.1771 \\
 & & & S2 Frozen (Prop.) & 0.3345 \(\pm\) 0.0243 & 0.5795 \(\pm\) 0.0202 & 41.0187 \(\pm\) 2.8090 & 5.0812 \(\pm\) 0.1668 \\
\cmidrule{3-8}
 & & \multirow{3}{*}{SPRN} 
     & S1 Only & 0.3150 \(\pm\) 0.0174 & \textbf{0.5652 \(\pm\) 0.0131} & 42.1829 \(\pm\) 1.8637 & 5.1875 \(\pm\) 0.1416 \\
 & & & S2 Unfrozen & 0.3141 \(\pm\) 0.0196 & 0.5646 \(\pm\) 0.0148 & 42.2364 \(\pm\) 1.9521 & 5.1913 \(\pm\) 0.1468 \\
 & & & S2 Frozen (Prop.) & \textbf{0.3155 \(\pm\) 0.0183} & 0.5651 \(\pm\) 0.0140 & \textbf{42.1484 \(\pm\) 1.9028} & \textbf{5.1841 \(\pm\) 0.1468} \\
\bottomrule
\multicolumn{8}{p{1.22\linewidth}}{\footnotesize \textbf{Note:} \textit{S1 Only}: Evaluated directly after Stage-1 opportunistic pretraining without adaptation. \textit{S2 Unfrozen}: Full-network fine-tuning on the common cohort. \textit{S2 Frozen}: Gradient updates were restricted to terminal-head parameters; batch-normalization running statistics continued to update during adaptation. \textit{S2 No \(L_2\)-SP}: Full-network adaptation without proximal regularization. \textit{S2 \(L_2\)-SP}: Full-network adaptation anchored by proximal regularization. \textit{(Prop.)} denotes the proposed default pipeline setting for the respective architecture.}
\end{tabular}
}
\end{table*}

\section{Discussion}
\label{sec:discussion}

\subsection{Learning from Partially Observed Training Cohorts}
The principal finding is that useful multimodal prediction did not require every training participant to have complete imaging. The complete opportunistic strategy improved mean \(R^2\) in most matched training comparisons, while prediction fusion improved all four reported mean metrics over the strongest constituent within each configuration. The results support separating the participants available for modality-specific fitting from those required to develop and evaluate a particular ensemble.

This contribution builds on opportunistic stacking for neuroimaging and cognitive prediction \cite{engemann2020combining, pat2022longitudinally}. Here, heterogeneous graph-based and dense predictors were trained independently, adapted using a common training cohort, and combined without a shared representation-learning objective. The comparatively large gains of the trimodal MAC-Net are consistent with the hypothesis that complete-case restrictions can limit modality-specific learning. However, the reported common-only versus two-stage comparisons estimate the benefit of the complete protocols and do not isolate the contribution of adaptation beyond pretraining.

Within-configuration fusion gains provide the clearest evidence that a single constituent did not exhaust the predictive value available from the fitted outputs. This is consistent with other multimodal stacking studies \cite{pat2022longitudinally, tetereva2025stacking}. Fusion may exploit complementary information, improve calibration, or more effectively combine correlated errors; the present comparisons do not distinguish among these mechanisms. Because each constituent also includes covariates, the ensemble gains cannot be attributed exclusively to complementary neural information.

\subsection{Mechanisms of Two-Stage Transfer: Pretraining and Adaptation}
The ablation analyses help clarify the relative contributions of expanded Stage 1 training and Stage 2 adaptation in the selected leading configurations. Models evaluated directly after Stage 1 generally retained performance close to that of the complete two-stage protocol. For example, the Stage 1 SPRN checkpoint for crystallized cognition achieved the same reported mean \(R^2\) as the complete protocol, whereas the Stage 1 MAC-Net checkpoint for total cognition had a mean \(R^2\) of 0.3348 compared with 0.3345 after adaptation. These findings indicate that expanded modality-specific training accounted for much of the observed difference from common-cohort-only training.

Stage 2 adaptation nevertheless produced different patterns across architectures. The dense-network ablations showed mixed effects of full-network versus head-only updating, without a consistent advantage for either adaptation strategy across targets and architectures. One interpretation is that updating a large feature extractor on the smaller common cohort may not provide a consistent advantage once a representation has been learned from the expanded Stage 1 pool. In this setting, head-only adaptation provided a conservative means of recalibrating the final prediction layer while retaining the broader Stage 1 representation.

IntelliGNN showed a different pattern. Among the tested variants, regularized full-network adaptation with \(L_2\)-SP achieved the highest observed mean \(R^2\) for the structural branch. In contrast, unregularized adaptation did not improve on the Stage 1 checkpoint, and head-only adaptation produced an intermediate result. These findings suggest that the graph-based model may benefit from limited adaptation of its relational representation, provided that parameter changes remain anchored to the Stage 1 solution. However, these results are specific to the evaluated architectures, targets, and training settings and should not be interpreted as establishing a general transfer-learning mechanism.

Across the selected ablations, differences between Stage-1-only checkpoints and the default two-stage protocol were small and did not consistently favor adaptation across metrics. These results do not establish that Stage 2 improves the ordering of predictions or is necessary for late fusion. Further evaluation using cross-fitted predictions and explicit calibration analyses would help assess its contribution.

\subsection{Target-Specific Modality Synergies and Metric Trade-offs}
\label{sec:discussion_target_synergies}

A critical finding of this methodological evaluation is that the optimal modality configuration is not universal, but rather inherently target-dependent. For instance, the bimodal sMRI+rsfMRI ensemble achieved the highest observed mean \(R^2\) (0.3706) and Pearson correlation (0.6108) for \(G_c\). For \(G_f\), tfMRI+rsfMRI achieved the highest observed mean \(R^2\) (0.2222) and the lowest MSE (68.6384) and MAE (6.5746), although its \(R^2\) was close to that of sMRI+rsfMRI (0.2216). A biologically plausible interpretation is that accumulated knowledge (\(G_c\)) primarily benefits from the joint information captured by stable morphometry and intrinsic baseline connectivity, whereas broader cognitive execution (\(G_f\)) draws heavily on the dynamic functional information captured during active task performance. However, because the present analysis validates a predictive pipeline rather than testing causal models, these modality-specific patterns are better interpreted as differences in the information emphasized by the fitted predictors than as evidence for distinct causal neurobiological mechanisms.


Across the configuration-specific cohorts, \(R^2\) and Pearson correlation did not always rank ensembles in the same order as MSE and MAE (Table \ref{tab:rerun_ensemble}). For \(G_c\), the trimodal ensemble had lower mean MSE and MAE but lower mean \(R^2\) and Pearson correlation than sMRI+rsfMRI. These results were obtained from different eligible cohorts and therefore do not isolate the effect of adding tfMRI. For \(G_f\), tfMRI+rsfMRI achieved a slightly higher mean \(R^2\) than sMRI+rsfMRI (0.2222 vs. 0.2216) and lower MSE and MAE, whereas sMRI+rsfMRI achieved a slightly higher Pearson correlation (0.4773 vs. 0.4743). These comparisons are descriptive because the configurations were evaluated in different modality-eligible cohorts.

The absence of a consistent trimodal advantage indicates that adding modalities did not uniformly improve performance in the evaluated cohorts. Modality redundancy, heightened estimation variability, and the significantly smaller size of the eligible trimodal cohort (\(N=4{,}216\) vs. \(N=6{,}845\) for sMRI+tfMRI) are highly plausible contributors to this phenomenon. Selection of an architectural configuration should therefore reflect the intended deployment setting and its specific error tolerance, rather than relying on a single cross-cohort maximum.

\subsection{Statistical Suppression and Biological Interpretation}
\label{sec:discussion_suppression}

The distribution of cross-validated Ridge coefficients challenges the assumption that late fusion merely accumulates positive predictive signals \cite{haufe2014interpretation}. In the trimodal ensembles, the mean sMRI coefficient was negative for \(G_c\) (-0.134) and \(G_t\) (-0.043), while the functional coefficients remained positive. These negative weights are compatible with suppression-like behavior among correlated predictors, but do not independently establish a specific error-correction mechanism. They describe how the fitted meta-regressor combined modality-specific predictions and should not be interpreted as evidence that structural information has a negative biological effect \cite{haufe2014interpretation}.


This systemic ensemble-level behavior must also be strictly separated from the overwhelming predominance of negative Integrated Gradients (IG) directions observed within the structural attribution summaries (Supplementary Table S6a). IG mathematically quantifies a geometric feature's contribution strictly relative to a zero-reference baseline along an integration path, rather than modeling a simple anatomical correlation or a localized derivative. The negative IG directions describe reference-relative contributions to the transformed model output and should not be interpreted as negative anatomical correlations with cognitive ability.


\subsection{Agreement and Differences Across XAI Methods}
\label{sec:discussion_xai_robustness}
IG and Input $\times$ Gradient produced highly correlated rankings of attribution magnitude, whereas raw Saliency differed substantially. Across all optimal configurations (\(G_f\), \(G_c\), and \(G_t\)), the convergence matrices revealed Pearson correlations of \(r \ge 0.9596\) between IG and Input $\times$ Gradient (IxG). These methods both incorporate input-relative scaling, but they quantify different quantities: IG integrates gradients along a reference-to-input path, whereas IxG multiplies the local gradient at the observed input by its value \cite{sundararajan2017axiomatic,samek2021explaining}. IG was designed to satisfy Sensitivity and Implementation Invariance; input scaling alone does not give IxG the same axiomatic guarantees. Raw Saliency instead quantifies local, unscaled gradient sensitivity and exhibited negligible or slightly negative correlations with the scaled methods (e.g., \(r = -0.3463\) against IG for \(G_c\)). 

This divergence shows that attribution rankings depend strongly on the attribution formulation and supports reporting method-specific sensitivity analyses. Sanity-check experiments show that some explanation methods can be insensitive to learned parameters or training labels; this finding does not imply that vanilla input-gradient Saliency generally fails these checks \cite{adebayo2018sanity}. The observed agreement between IG and IxG therefore describes consistency between two related gradient-based summaries, rather than establishing explanation fidelity or biological validity. Such interpretation requires additional evaluation, consistent with the broader medical-imaging XAI literature \cite{vanderVelden2022explainable}.

\subsection{Acquisition Context and Technical Confounders}
\label{sec:discussion_covariates}
ABCD imaging measures can retain acquisition-related differences despite standardized protocols and centralized processing. Hagler et al. documented differences across scanner manufacturers in structural, diffusion, task-based, and resting-state measures and recommended including the unique scanner device serial number as a categorical covariate in group analyses, or applying statistical harmonization \cite{hagler2019image}. Including one-hot-encoded scanner identity in each modality-specific predictor followed this recommendation's rationale by allowing predictions to depend on acquisition context alongside demographic and motion covariates.

Scanner identity received the highest normalized grouped composite covariate score across the evaluated configurations. These scores averaged the normalized IG, Saliency, and Input $\times$ Gradient magnitudes, summed the scores of indicators for each categorical variable, and normalized the resulting grouped values. The maximum score of 1.00, therefore, denotes the highest grouped attribution within a summary, rather than a proportion of predictive variance. Grouped scores can also depend on the number of indicator columns, so scanner identity's ranking should be interpreted with this aggregation in mind.

The inclusion of scanner covariates is supported by ABCD guidance, but the present analysis does not establish that scanner effects were removed or that imaging representations became independent of confounders \cite{hagler2019image,rao2017predictive,snoek2019how}. Scanner attribution describes the sensitivity of fitted predictions and cannot separate technical acquisition effects from site-associated participant characteristics or correlated predictors. Future evaluation should compare alternative covariate handling and explicit harmonization, and test generalization to held-out sites or scanners \cite{dudley2023harmonizer}. Any learned preprocessing or harmonization parameters should be estimated within the relevant training partition to avoid leakage \cite{rosenblatt2024leakage}.

\subsection{Target-Dependent Modality Weighting}
\label{sec:discussion_modality_dominance}
The final RidgeCV fusion weights showed target-dependent reliance on the fitted modality-specific predictions. For \(G_f\), the mean Ridge coefficient was 0.68 for tfMRI and 0.40 for rsfMRI, while attribution summaries highlighted task-evoked features including the left caudal middle frontal gyrus. This pattern is consistent with previous evidence that task-induced brain states can improve prediction of individual traits \cite{greene2018task}. However, coefficients of constituent predictions do not establish which individual brain regions caused the fusion gain or demonstrate a specific cognitive mechanism \cite{haufe2014interpretation}.

For \(G_c\), the optimal structural--resting-state fusion placed a larger mean coefficient on the rsfMRI prediction (0.74) than on the sMRI prediction (0.37). This describes the relative predictive contribution of the fitted outputs in that cohort; it does not establish that accumulated knowledge is primarily determined by intrinsic functional architecture or morphology. Comparative multimodal studies show that predictive performance depends on the phenotype and modality configuration, while reviews emphasize the methodological dependence of structure--function--cognition associations \cite{ooi2022comparison,tetereva2025stacking,litwinczuk2023relating}.



The rsfMRI findings similarly favored a complex, graded network integration over a simplistic attention-versus-default-mode dichotomy. While dorsal-attention connectivity was prominent in the \(G_f\) tfMRI+rsfMRI pipeline, left default-network regions ranked highly in the \(G_c\) attribution summaries. However, the \(G_c\) sMRI+rsfMRI architecture concurrently emphasized ventral-attention and sensory connections, indicating that the most reproducible predictive pattern was regional involvement mapping across distributed functional systems, rather than isolated intra-network dominance \cite{bassett2017network}.

\subsection{Translational Implications}
The immediate contribution of this study is methodological. Participants with incomplete modality coverage can contribute to modality-specific model development while ensemble development and evaluation remain defined on a common cohort. This approach directly addresses the pervasive ``missing modality'' problem encountered in real-world clinical workflows and large-scale research datasets, where imaging availability often varies due to scan-time constraints, motion artifacts, or patient non-compliance. It does not remove the need for the modalities specified by a fitted ensemble at inference, nor does it establish that incomplete coverage is random or that expanded training eliminates selection bias.

The present models predict contemporaneous baseline cognitive scores and have not been evaluated for diagnosis, prognosis, treatment selection, or replacement of standardized cognitive assessment. Nevertheless, if the framework is externally validated in clinically relevant populations, multimodal MRI-based cognitive prediction could potentially support MRI-informed phenotyping, stratification of participants for research studies, or identification of individuals who may benefit from comprehensive cognitive assessment. Such applications may be particularly relevant when standardized testing is incomplete, impractical, or affected by fatigue, developmental stage, communication difficulty, or clinical impairment.

Any future clinical role should be adjunctive rather than substitutive. The added value of MRI-derived predictions would need to be demonstrated against routinely available demographic, clinical, neuropsychological, and imaging information. Translation would require a clearly defined use case, external validation across populations and acquisition settings, calibration and subgroup analyses, assessment of fairness and acquisition dependence, and evidence that model use improves the relevant clinical decision pathway \cite{collins2024tripodai,vasey2022decideai}. Prospective evaluation would be necessary before considering clinical decision-support applications.

\subsection{Limitations and Future Evaluation}
Several limitations qualify these findings. First, the reported means and standard deviations describe variation across outer folds and do not establish statistically significant differences or equivalence between leading configurations. The small differences among some configurations should therefore be interpreted descriptively. Family-aware uncertainty estimates and comparisons that account for dependence induced by resampling would strengthen future evaluations.

Second, the same validation partition was used for neural checkpoint selection and RidgeCV fusion fitting. The fusion procedure, therefore, represents validation-set blending rather than fully cross-fitted stacking. In addition, covariate-only models were refitted on training and validation participants after tuning, whereas neural models used selected checkpoints. The reported comparisons should consequently be interpreted as comparisons of complete fitting protocols rather than as isolated estimates of individual training components.

Third, preprocessing for rsfMRI feature-selection tuning was fitted once to the selection-training pool rather than refitted within each internal split. Outer-test participants remained excluded, but this internal procedure was not fully nested. The categorical vocabulary also used outcome-independent covariate values from the configuration cohort. Future studies should evaluate strictly training-derived category handling, including scenarios with unseen scanners.

Fourth, each modality configuration was evaluated in a distinct eligible cohort. Neither the observed cross-configuration ranking nor the expanded Stage 1 pools establishes that missingness was random or that the models generalize to participants excluded by imaging-quality criteria. Matched-participant modality comparisons and explicit analyses of eligibility-related selection would help separate modality effects from cohort composition.

Finally, a key limitation of the present study is its exclusive reliance on cross-sectional baseline assessments. While evaluating baseline data was essential to initially establish the validity of the opportunistic pretraining and multimodal fusion framework, it does not account for the longitudinal nature of the ABCD study. Brain-behavior relationships are highly dynamic, particularly during adolescence, and examining cross-sectional data alone misses an opportunity to track developmental changes. Future research must prioritize adapting these deep ensemble architectures to incorporate longitudinal neuroimaging and behavioral data, enabling the robust modeling of within-person trajectories. Extending this framework to capture longitudinal changes will be critical for understanding how multimodal predictions evolve over time and for eventually evaluating their real-world clinical utility.

\section{Conclusion}
\label{sec:conclusion}
This study evaluated a two-stage, prediction-level ensemble framework for predicting cognitive scores from incomplete multimodal MRI data. The framework used expanded modality-specific training pools while preserving family-aware evaluation on the appropriate complete-data cohort. In 24 of 27 matched comparisons, the complete two-stage protocol achieved a higher mean \(R^2\) than common-cohort-only training, and late fusion improved the reported mean performance metrics within each multimodal configuration.

The ablation analyses indicated that expanded Stage~1 pretraining accounted for much of the observed performance difference, whereas the relative value of Stage~2 adaptation depended on the architecture. Head-only adaptation performed competitively for the evaluated dense functional models, while regularized full-network adaptation achieved the highest observed structural-model performance. These findings support architecture-specific evaluation of adaptation strategies when modality-specific models are trained on partially overlapping cohorts.

Attribution analyses identified target-dependent features emphasized by the fitted models, including recurring fusiform structural regions, N-back task features for fluid and total cognition, and distributed resting-state connectivity patterns. Scanner identity also had a substantial covariate attribution across models, highlighting the importance of acquisition context. These findings describe model behavior and should not be interpreted as causal neurobiological mechanisms.

By decoupling initial representation learning from intersection-based fusion, the framework enables researchers to leverage modality-specific observations that would otherwise be excluded by strict complete-case requirements, offering a scalable solution for incomplete multimodal datasets. With external validation in patient populations and across acquisition settings, MRI-informed cognitive prediction could eventually support cognitive phenotyping, research stratification, or adjunctive assessment when standardized testing is incomplete or difficult to obtain. However, the present results do not establish clinical utility. Prospective validation, calibration assessment, evaluation of subgroup and acquisition-related performance, and evidence of benefit within a defined clinical pathway are required before clinical implementation.

\section*{Acknowledgment}
This research was supported by the Science and Technology Fellowship Trust, Government of the People's Republic of Bangladesh, and the Commonwealth through an Australian Government Research Training Program Scholarship. Data used in the preparation of this article were obtained from the Adolescent Brain Cognitive Development (ABCD) Study Data Release 5.1 (http://dx.doi.org/10.15154/z563-zd24), held in the NIH Brain Development Cohorts Data Sharing Platform. In addition, this work was supported by resources provided by The University of Queensland Research Computing Center’s Bunya supercomputer.

\printcredits
\bibliographystyle{elsarticle-num}
\bibliography{cas-refs}

\clearpage
\onecolumn
\raggedbottom

\pagestyle{fancy}
\fancyhf{} 
\renewcommand{\headrulewidth}{0pt} 
\cfoot{\footnotesize Page \thepage\ of \pageref{LastPage}}

\raggedbottom

\begin{center}
\textbf{\large Supplementary Materials}
\end{center}

\vspace{.1em}

\begin{center}
\textbf{\large Brain–Behavior Modeling from Multimodal MRI using Opportunistic Pretraining and Common-Cohort Adaptation}
\end{center}

\setcounter{table}{0}

\renewcommand{\thetable}{S1}
\begingroup\footnotesize
\begin{longtable}{@{\extracolsep{\fill}}llp{11cm}@{}}
\caption{Brain regions and parcellations utilized within each modality-specific architecture.}\\
\toprule
Architecture & Atlas & Regions\\\midrule\endfirsthead
\multicolumn{3}{l}{\tablename\ \thetable\ (continued)}\\
\toprule
Architecture & Atlas & Regions\\\midrule\endhead
\bottomrule\endfoot
IntelliGNN \& MAC-Net & Desikan-Killiany & \textbf{Cortical (68 nodes, Bilateral):} Banks Superior Temporal Sulcus, Caudal Anterior Cingulate, Caudal Middle Frontal, Cuneus, Entorhinal, Fusiform, Inferior Parietal, Inferior Temporal, Isthmus Cingulate, Lateral Occipital, Lateral Orbitofrontal, Lingual, Medial Orbitofrontal, Middle Temporal, Parahippocampal, Paracentral, Pars Opercularis, Pars Orbitalis, Pars Triangularis, Pericalcarine, Postcentral, Posterior Cingulate, Precentral, Precuneus, Rostral Anterior Cingulate, Rostral Middle Frontal, Superior Frontal, Superior Parietal, Superior Temporal, Supramarginal, Frontal Pole, Temporal Pole, Transverse Temporal, Insula.\\\addlinespace
& ASEG & \textbf{Subcortical (26 nodes, Bilateral):} Cerebral White Matter, Lateral Ventricle, Inferior Lateral Ventricle, Cerebellar White Matter, Cerebellar Cortex, Thalamus Proper, Caudate, Putamen, Pallidum, Hippocampus, Amygdala, Accumbens Area, Ventral Diencephalon.\\\midrule
SPRN & Gordon 333 & \textbf{Cortical Networks (333 nodes):} Default, Somatomotor (Hand and Mouth), Visual, Fronto-Parietal, Auditory, Cingulo-Parietal, Retrosplenial Temporal, Cingulo-Opercular, Ventral Attention, Dorsal Attention, Salience, Unassigned.\\\addlinespace
& ASEG & \textbf{Subcortical (19 nodes):} \textbf{Bilateral:} Cerebellum, Thalamus, Caudate, Putamen, Pallidum, Hippocampus, Amygdala, Accumbens, Ventral Diencephalon. \textbf{Central:} Brain Stem.\\
\end{longtable}
\endgroup

\renewcommand{\thetable}{S2}
\begingroup\footnotesize
\begin{longtable}{@{\extracolsep{\fill}}llp{12cm}@{}}
\caption{Task conditions and behavioral contrasts provided in exact model input order for MAC-Net (tfMRI).}\\
\toprule
Task & Category & Conditions / Contrasts\\\midrule\endfirsthead
\multicolumn{3}{l}{\tablename\ \thetable\ (continued)}\\
\toprule
Task & Category & Conditions / Contrasts\\\midrule\endhead
\bottomrule\endfoot
N-back & Conditions & (1) 0-back, (2) 2-back, (3) Place, (4) Emotion.\\\addlinespace
 & Contrasts & (5) 2-back vs 0-back, (6) Face vs Place, (7) Emotion vs Neutral, (8) Negative vs Neutral, (9) Positive vs Neutral.\\\midrule
MID & Contrasts & (10) Anticipation Large vs Small Loss, (11) Anticipation Loss vs Neutral, (12) Anticipation Large Reward vs Neutral, (13) Anticipation Large Loss vs Neutral, (14) Reward Positive vs Negative Feedback, (15) Loss Positive vs Negative Feedback, (16) Anticipation Small Reward vs Neutral, (17) Anticipation Small Loss vs Neutral, (18) Anticipation Reward vs Neutral, (19) Anticipation Large vs Small Reward.\\
\end{longtable}
\endgroup

\renewcommand{\thetable}{S3a}
\begingroup\footnotesize
\begin{longtable}{@{\extracolsep{\fill}}llllll@{}}
\caption{Covariate-only baseline predictive performance for crystallized cognition (\(G_c\)).}\\
\toprule
Cohort & Model & \(R^2\) & \(r\) & MSE & MAE\\\midrule\endfirsthead
\multicolumn{6}{l}{\tablename\ \thetable\ (continued)}\\\toprule
Cohort & Model & \(R^2\) & \(r\) & MSE & MAE\\\midrule\endhead\bottomrule\endfoot
S & Ridge & 0.3020 \(\pm\) 0.0146 & 0.5507 \(\pm\) 0.0140 & 33.8094 \(\pm\) 0.7981 & 4.5102 \(\pm\) 0.0575\\
S & Lasso & 0.3015 \(\pm\) 0.0162 & 0.5504 \(\pm\) 0.0154 & 33.8346 \(\pm\) 0.8892 & 4.5128 \(\pm\) 0.0603\\
S & Elastic Net & 0.2983 \(\pm\) 0.0155 & 0.5482 \(\pm\) 0.0151 & 33.9870 \(\pm\) 0.8115 & 4.5235 \(\pm\) 0.0549\\
S & RF & 0.2819 \(\pm\) 0.0171 & 0.5317 \(\pm\) 0.0162 & 34.7746 \(\pm\) 0.5975 & 4.5808 \(\pm\) 0.0377\\
S & MLP & 0.3012 \(\pm\) 0.0128 & 0.5506 \(\pm\) 0.0126 & 33.8519 \(\pm\) 0.8714 & 4.5121 \(\pm\) 0.0479\\
T & Ridge & 0.2619 \(\pm\) 0.0116 & 0.5145 \(\pm\) 0.0125 & 31.6688 \(\pm\) 1.4650 & 4.3692 \(\pm\) 0.0863\\
T & Lasso & 0.2625 \(\pm\) 0.0109 & 0.5153 \(\pm\) 0.0118 & 31.6428 \(\pm\) 1.4800 & 4.3685 \(\pm\) 0.0838\\
T & Elastic Net & 0.2590 \(\pm\) 0.0089 & 0.5124 \(\pm\) 0.0106 & 31.8015 \(\pm\) 1.6204 & 4.3767 \(\pm\) 0.1011\\
\end{longtable}
\endgroup

\renewcommand{\thetable}{S3b}
\begingroup\footnotesize
\begin{longtable}{@{\extracolsep{\fill}}llllll@{}}
\caption{Covariate-only baseline predictive performance for fluid cognition (\(G_f\)).}\\
\toprule
Cohort & Model & \(R^2\) & \(r\) & MSE & MAE\\\midrule\endfirsthead
\multicolumn{6}{l}{\tablename\ \thetable\ (continued)}\\\toprule
Cohort & Model & \(R^2\) & \(r\) & MSE & MAE\\\midrule\endhead\bottomrule\endfoot
S & Ridge & 0.1899 \(\pm\) 0.0153 & 0.4390 \(\pm\) 0.0147 & 88.2052 \(\pm\) 2.4446 & 7.3982 \(\pm\) 0.1028\\
S & Lasso & 0.1888 \(\pm\) 0.0144 & 0.4379 \(\pm\) 0.0138 & 88.3249 \(\pm\) 2.4258 & 7.4017 \(\pm\) 0.0972\\
S & Elastic Net & 0.1854 \(\pm\) 0.0126 & 0.4343 \(\pm\) 0.0121 & 88.7022 \(\pm\) 2.3919 & 7.4180 \(\pm\) 0.0962\\
S & RF & 0.1686 \(\pm\) 0.0089 & 0.4145 \(\pm\) 0.0072 & 90.5318 \(\pm\) 2.2640 & 7.5029 \(\pm\) 0.1030\\
S & MLP & 0.1857 \(\pm\) 0.0144 & 0.4354 \(\pm\) 0.0125 & 88.6616 \(\pm\) 2.2901 & 7.4264 \(\pm\) 0.1027\\
T & Ridge & 0.1535 \(\pm\) 0.0053 & 0.3943 \(\pm\) 0.0074 & 77.5943 \(\pm\) 3.6491 & 6.9554 \(\pm\) 0.1677\\
T & Lasso & 0.1522 \(\pm\) 0.0055 & 0.3927 \(\pm\) 0.0082 & 77.7060 \(\pm\) 3.5698 & 6.9614 \(\pm\) 0.1658\\
T & Elastic Net & 0.1509 \(\pm\) 0.0043 & 0.3912 \(\pm\) 0.0071 & 77.8262 \(\pm\) 3.5837 & 6.9679 \(\pm\) 0.1647\\
\end{longtable}
\endgroup

\renewcommand{\thetable}{S3c}
\begingroup\footnotesize
\begin{longtable}{@{\extracolsep{\fill}}llllll@{}}
\caption{Covariate-only baseline predictive performance for total cognition (\(G_t\)).}\\
\toprule
Cohort & Model & \(R^2\) & \(r\) & MSE & MAE\\\midrule\endfirsthead
\multicolumn{6}{l}{\tablename\ \thetable\ (continued)}\\\toprule
Cohort & Model & \(R^2\) & \(r\) & MSE & MAE\\\midrule\endhead\bottomrule\endfoot
S & Ridge & 0.3016 \(\pm\) 0.0110 & 0.5510 \(\pm\) 0.0087 & 55.8309 \(\pm\) 1.7076 & 5.8780 \(\pm\) 0.0782\\
S & Lasso & 0.3009 \(\pm\) 0.0113 & 0.5504 \(\pm\) 0.0088 & 55.8858 \(\pm\) 1.7216 & 5.8813 \(\pm\) 0.0742\\
S & Elastic Net & 0.2949 \(\pm\) 0.0101 & 0.5453 \(\pm\) 0.0081 & 56.3657 \(\pm\) 1.6291 & 5.9041 \(\pm\) 0.0694\\
S & RF & 0.2784 \(\pm\) 0.0128 & 0.5297 \(\pm\) 0.0120 & 57.6646 \(\pm\) 0.8868 & 5.9743 \(\pm\) 0.0464\\
S & MLP & 0.2986 \(\pm\) 0.0130 & 0.5488 \(\pm\) 0.0100 & 56.0791 \(\pm\) 2.0593 & 5.9009 \(\pm\) 0.1016\\
T & Ridge & 0.2620 \(\pm\) 0.0117 & 0.5131 \(\pm\) 0.0116 & 48.6221 \(\pm\) 2.0294 & 5.5149 \(\pm\) 0.1286\\
T & Lasso & 0.2646 \(\pm\) 0.0108 & 0.5154 \(\pm\) 0.0103 & 48.4534 \(\pm\) 2.0094 & 5.5084 \(\pm\) 0.1291\\
T & Elastic Net & 0.2615 \(\pm\) 0.0133 & 0.5127 \(\pm\) 0.0132 & 48.6575 \(\pm\) 2.1879 & 5.5127 \(\pm\) 0.1417\\
\end{longtable}
\endgroup

\renewcommand{\thetable}{S4a}
\begingroup\footnotesize
\begin{longtable}{@{\extracolsep{\fill}}lllllll@{}}
\caption{Brain-plus-covariate predictive performance for crystallized cognition (\(G_c\)). U: unimodal; C: common-cohort-only training; TS: two-stage opportunistic training.}\\
\toprule
Cohort & Model & Strategy & \(R^2\) & \(r\) & MSE & MAE\\\midrule\endfirsthead
\multicolumn{7}{l}{\tablename\ \thetable\ (continued)}\\\toprule
Cohort & Model & Strategy & \(R^2\) & \(r\) & MSE & MAE\\\midrule\endhead\bottomrule\endfoot
S & IntelliGNN & U & 0.3166 \(\pm\) 0.0176 & 0.5635 \(\pm\) 0.0162 & 33.1018 \(\pm\) 0.8867 & 4.4649 \(\pm\) 0.0607\\
S+R & IntelliGNN & C & 0.2801 \(\pm\) 0.0139 & 0.5330 \(\pm\) 0.0127 & 31.6613 \(\pm\) 1.3799 & 4.3985 \(\pm\) 0.1109\\
S+R & IntelliGNN & TS & 0.2934 \(\pm\) 0.0140 & 0.5427 \(\pm\) 0.0128 & 31.0790 \(\pm\) 1.4264 & 4.3582 \(\pm\) 0.1035\\
S+T & IntelliGNN & C & 0.2596 \(\pm\) 0.0101 & 0.5111 \(\pm\) 0.0106 & 31.6634 \(\pm\) 1.2773 & 4.3706 \(\pm\) 0.0722\\
S+T & IntelliGNN & TS & 0.2620 \(\pm\) 0.0130 & 0.5148 \(\pm\) 0.0113 & 31.5654 \(\pm\) 1.4542 & 4.3659 \(\pm\) 0.0968\\
S+T+R & IntelliGNN & C & 0.2412 \(\pm\) 0.0111 & 0.4937 \(\pm\) 0.0111 & 30.2460 \(\pm\) 0.8376 & 4.2986 \(\pm\) 0.0775\\
S+T+R & IntelliGNN & TS & 0.2520 \(\pm\) 0.0205 & 0.5068 \(\pm\) 0.0170 & 29.8008 \(\pm\) 0.6651 & 4.2650 \(\pm\) 0.0589\\
T & MAC-Net & U & 0.3114 \(\pm\) 0.0113 & 0.5591 \(\pm\) 0.0103 & 29.5441 \(\pm\) 1.3299 & 4.2313 \(\pm\) 0.0841\\
\end{longtable}
\endgroup

\renewcommand{\thetable}{S4b}
\begingroup\footnotesize
\begin{longtable}{@{\extracolsep{\fill}}lllllll@{}}
\caption{Brain-plus-covariate predictive performance for fluid cognition (\(G_f\)). U: unimodal; C: common-cohort-only training; TS: two-stage opportunistic training.}\\
\toprule
Cohort & Model & Strategy & \(R^2\) & \(r\) & MSE & MAE\\\midrule\endfirsthead
\multicolumn{7}{l}{\tablename\ \thetable\ (continued)}\\\toprule
Cohort & Model & Strategy & \(R^2\) & \(r\) & MSE & MAE\\\midrule\endhead\bottomrule\endfoot
S & IntelliGNN & U & 0.2023 \(\pm\) 0.0153 & 0.4530 \(\pm\) 0.0134 & 86.8566 \(\pm\) 2.3671 & 7.3535 \(\pm\) 0.0929\\
S+R & IntelliGNN & C & 0.1741 \(\pm\) 0.0154 & 0.4217 \(\pm\) 0.0176 & 83.1039 \(\pm\) 2.7670 & 7.1735 \(\pm\) 0.1629\\
S+R & IntelliGNN & TS & 0.1822 \(\pm\) 0.0150 & 0.4288 \(\pm\) 0.0178 & 82.3039 \(\pm\) 3.0911 & 7.1439 \(\pm\) 0.1790\\
S+T & IntelliGNN & C & 0.1517 \(\pm\) 0.0207 & 0.3905 \(\pm\) 0.0265 & 77.0592 \(\pm\) 3.9948 & 6.9630 \(\pm\) 0.1751\\
S+T & IntelliGNN & TS & 0.1551 \(\pm\) 0.0209 & 0.3968 \(\pm\) 0.0252 & 76.7451 \(\pm\) 3.9991 & 6.9424 \(\pm\) 0.1693\\
S+T+R & IntelliGNN & C & 0.1369 \(\pm\) 0.0127 & 0.3740 \(\pm\) 0.0133 & 76.1356 \(\pm\) 3.2643 & 6.9173 \(\pm\) 0.1685\\
S+T+R & IntelliGNN & TS & 0.1381 \(\pm\) 0.0214 & 0.3825 \(\pm\) 0.0200 & 76.0061 \(\pm\) 3.0066 & 6.8984 \(\pm\) 0.1435\\
T & MAC-Net & U & 0.2007 \(\pm\) 0.0208 & 0.4535 \(\pm\) 0.0186 & 73.2790 \(\pm\) 4.1989 & 6.7683 \(\pm\) 0.2165\\
\end{longtable}
\endgroup

\renewcommand{\thetable}{S4c}
\begingroup\footnotesize
\begin{longtable}{@{\extracolsep{\fill}}lllllll@{}}
\caption{Brain-plus-covariate predictive performance for total cognition (\(G_t\)). U: unimodal; C: common-cohort-only training; TS: two-stage opportunistic training.}\\
\toprule
Cohort & Model & Strategy & \(R^2\) & \(r\) & MSE & MAE\\\midrule\endfirsthead
\multicolumn{7}{l}{\tablename\ \thetable\ (continued)}\\\toprule
Cohort & Model & Strategy & \(R^2\) & \(r\) & MSE & MAE\\\midrule\endhead\bottomrule\endfoot
S & IntelliGNN & U & 0.3137 \(\pm\) 0.0130 & 0.5622 \(\pm\) 0.0102 & 54.8532 \(\pm\) 1.3464 & 5.8370 \(\pm\) 0.0694\\
S+R & IntelliGNN & C & 0.2861 \(\pm\) 0.0233 & 0.5371 \(\pm\) 0.0213 & 51.1280 \(\pm\) 2.1220 & 5.6338 \(\pm\) 0.1220\\
S+R & IntelliGNN & TS & 0.2950 \(\pm\) 0.0241 & 0.5449 \(\pm\) 0.0226 & 50.4800 \(\pm\) 1.7229 & 5.6106 \(\pm\) 0.0839\\
S+T & IntelliGNN & C & 0.2565 \(\pm\) 0.0245 & 0.5074 \(\pm\) 0.0244 & 48.4818 \(\pm\) 2.2729 & 5.5119 \(\pm\) 0.1070\\
S+T & IntelliGNN & TS & 0.2695 \(\pm\) 0.0252 & 0.5201 \(\pm\) 0.0237 & 47.6388 \(\pm\) 2.3303 & 5.4603 \(\pm\) 0.1111\\
S+T+R & IntelliGNN & C & 0.2498 \(\pm\) 0.0122 & 0.5021 \(\pm\) 0.0119 & 45.9180 \(\pm\) 2.4034 & 5.3753 \(\pm\) 0.1149\\
S+T+R & IntelliGNN & TS & 0.2505 \(\pm\) 0.0189 & 0.5075 \(\pm\) 0.0154 & 45.8621 \(\pm\) 2.3207 & 5.3630 \(\pm\) 0.1090\\
T & MAC-Net & U & 0.3353 \(\pm\) 0.0242 & 0.5811 \(\pm\) 0.0190 & 43.7696 \(\pm\) 1.8478 & 5.2277 \(\pm\) 0.1354\\
\end{longtable}
\endgroup

\renewcommand{\thetable}{S5}
\begingroup\footnotesize
\begin{longtable}{@{\extracolsep{\fill}}lllllll@{}}
\caption{Grouped covariate importance rankings across models and targets. Directions (signs) denote the sums of raw mean Integrated Gradients (IG); overall group magnitude scales with categorical count.}\\
\toprule
Target & Cohort & Model & Rank & Covariate & Sign & Score\\\midrule\endfirsthead
\multicolumn{7}{l}{\tablename\ \thetable\ (continued)}\\\toprule
Target & Cohort & Model & Rank & Covariate & Sign & Score\\\midrule\endhead\bottomrule\endfoot
Cryst & T+R & MACNET & 1 & MRI Device Serial Number & - & 1.0000\\
Cryst & T+R & MACNET & 2 & Sex & - & 0.1363\\
Cryst & T+R & MACNET & 3 & Age & + & 0.1199\\
Cryst & T+R & MACNET & 4 & Highest Parental Education & + & 0.0872\\
Cryst & T+R & MACNET & 5 & Total Intracranial Brain Volume & + & 0.0545\\
Cryst & T+R & MACNET & 6 & Combined Family Income & + & 0.0520\\
Cryst & T+R & MACNET & 7 & MID Task Mean Motion & + & 0.0373\\
Cryst & T+R & MACNET & 8 & N-Back Task Mean Motion & + & 0.0208\\
\end{longtable}
\endgroup

\renewcommand{\thetable}{S6a}
\begingroup\footnotesize
\begin{longtable}{@{\extracolsep{\fill}}lllllll@{}}
\caption{Top 10 specific structural and diffusion-derived (sMRI) features ranked by composite importance score.}\\
\toprule
Target & Cohort & Model & Rank & Feature & Sign & Score\\\midrule\endfirsthead
\multicolumn{7}{l}{\tablename\ \thetable\ (continued)}\\\toprule
Target & Cohort & Model & Rank & Feature & Sign & Score\\\midrule\endhead\bottomrule\endfoot
Cryst & S+R & INTELLIGNN & 1 & L. Fusiform [Sulcal Depth] & - & 1.0000\\
Cryst & S+R & INTELLIGNN & 2 & L. Fusiform [T2 Gray/White Contrast] & + & 0.7050\\
Cryst & S+R & INTELLIGNN & 3 & L. Fusiform [Volume] & - & 0.5475\\
Cryst & S+R & INTELLIGNN & 4 & R. Fusiform [Sulcal Depth] & - & 0.4257\\
Cryst & S+R & INTELLIGNN & 5 & L. Cerebellar Cortex [Volume] & - & 0.3824\\
Cryst & S+R & INTELLIGNN & 6 & L. Fusiform [T1 White] & - & 0.3459\\
Cryst & S+R & INTELLIGNN & 7 & L. Cerebellar White Matter [Volume] & - & 0.3454\\
Cryst & S+R & INTELLIGNN & 8 & R. Cerebellar Cortex [Volume] & - & 0.3298\\
\end{longtable}
\endgroup

\renewcommand{\thetable}{S6b}
\begingroup\footnotesize
\begin{longtable}{@{\extracolsep{\fill}}lllllll@{}}
\caption{Top 10 structural (sMRI) nodes, aggregated across multiple regional morphometric and microstructural features.}\\
\toprule
Target & Cohort & Model & Rank & Node & Sign & Score\\\midrule\endfirsthead
\multicolumn{7}{l}{\tablename\ \thetable\ (continued)}\\\toprule
Target & Cohort & Model & Rank & Node & Sign & Score\\\midrule\endhead\bottomrule\endfoot
Cryst & S+R & INTELLIGNN & 1 & L. Fusiform & - & 3.3790\\
Cryst & S+R & INTELLIGNN & 2 & R. Fusiform & - & 1.6270\\
Cryst & S+R & INTELLIGNN & 3 & L. Superior Temporal & - & 0.8183\\
Cryst & S+R & INTELLIGNN & 4 & R. Inferior Temporal & - & 0.6921\\
Cryst & S+R & INTELLIGNN & 5 & R. Superior Temporal & - & 0.6838\\
Cryst & S+R & INTELLIGNN & 6 & L. Inferior Temporal & - & 0.6415\\
Cryst & S+R & INTELLIGNN & 7 & R. Entorhinal & - & 0.6300\\
Cryst & S+R & INTELLIGNN & 8 & L. Cerebellar Cortex & - & 0.4957\\
\end{longtable}
\endgroup

\renewcommand{\thetable}{S7a}
\begingroup\footnotesize
\begin{longtable}{@{\extracolsep{\fill}}lllllll@{}}
\caption{Top 10 global resting-state functional connectivity (rsfMRI) edges (highways) ranked by composite importance score.}\\
\toprule
Target & Cohort & Model & Rank & Edge & Sign & Score\\\midrule\endfirsthead
\multicolumn{7}{l}{\tablename\ \thetable\ (continued)}\\\toprule
Target & Cohort & Model & Rank & Edge & Sign & Score\\\midrule\endhead\bottomrule\endfoot
Cryst & T+R & SPRN & 1 & R. Visual \(\leftrightarrow\) R. Retrosplenial Temporal & + & 0.8647\\
Cryst & T+R & SPRN & 2 & L. SM Hand \(\leftrightarrow\) L. None & - & 0.7789\\
Cryst & T+R & SPRN & 3 & R. Ventral Attention \(\leftrightarrow\) R. Visual & + & 0.7747\\
Cryst & T+R & SPRN & 4 & R. None \(\leftrightarrow\) R. Retrosplenial Temporal & + & 0.7502\\
Cryst & T+R & SPRN & 5 & L. SM Hand \(\leftrightarrow\) R. SM Hand & - & 0.7348\\
Cryst & T+R & SPRN & 6 & L. Default \(\leftrightarrow\) L. Default & + & 0.7235\\
Cryst & T+R & SPRN & 7 & L. Default \(\leftrightarrow\) R. Default & + & 0.7208\\
Cryst & T+R & SPRN & 8 & L. Default \(\leftrightarrow\) R. Dorsal Attention & + & 0.7010\\
\end{longtable}
\endgroup

\renewcommand{\thetable}{S7b}
\begingroup\footnotesize
\begin{longtable}{@{\extracolsep{\fill}}lllllll@{}}
\caption{Top 10 resting-state (rsfMRI) nodes (hubs), aggregated across all incident connectivity edges and regional variance.}\\
\toprule
Target & Cohort & Model & Rank & Node & Sign & Score\\\midrule\endfirsthead
\multicolumn{7}{l}{\tablename\ \thetable\ (continued)}\\\toprule
Target & Cohort & Model & Rank & Node & Sign & Score\\\midrule\endhead\bottomrule\endfoot
Cryst & T+R & SPRN & 1 & L. Default & + & 15.5405\\
Cryst & T+R & SPRN & 2 & L. Ventral Attention & - & 14.4449\\
Cryst & T+R & SPRN & 3 & R. Visual & + & 13.3295\\
Cryst & T+R & SPRN & 4 & L. Dorsal Attention & - & 12.7716\\
Cryst & T+R & SPRN & 5 & R. Default & + & 11.8552\\
Cryst & T+R & SPRN & 6 & L. Cingulo-Parietal & + & 11.4609\\
Cryst & T+R & SPRN & 7 & L. Cingulo-Parietal & + & 11.0881\\
Cryst & T+R & SPRN & 8 & L. Ventral Diencephalon & + & 10.7644\\
\end{longtable}
\endgroup

\renewcommand{\thetable}{S7c}
\begingroup\footnotesize
\begin{longtable}{@{\extracolsep{\fill}}lllllll@{}}
\caption{Top 10 specific resting-state (rsfMRI) features, including individual functional connectivity edges and regional signal variance.}\\
\toprule
Target & Cohort & Model & Rank & Feature & Sign & Score\\\midrule\endfirsthead
\multicolumn{7}{l}{\tablename\ \thetable\ (continued)}\\\toprule
Target & Cohort & Model & Rank & Feature & Sign & Score\\\midrule\endhead\bottomrule\endfoot
Cryst & T+R & SPRN & 1 & FC Edge: R. Visual \(\leftrightarrow\) R. Retrosplenial Temporal & + & 0.8647\\
Cryst & T+R & SPRN & 2 & FC Edge: L. SM Hand \(\leftrightarrow\) L. None & - & 0.7789\\
Cryst & T+R & SPRN & 3 & FC Edge: R. Ventral Attention \(\leftrightarrow\) R. Visual & + & 0.7747\\
Cryst & T+R & SPRN & 4 & FC Edge: R. None \(\leftrightarrow\) R. Retrosplenial Temporal & + & 0.7502\\
Cryst & T+R & SPRN & 5 & FC Edge: L. SM Hand \(\leftrightarrow\) R. SM Hand & - & 0.7348\\
Cryst & T+R & SPRN & 6 & FC Edge: L. Default \(\leftrightarrow\) L. Default & + & 0.7235\\
Cryst & T+R & SPRN & 7 & FC Edge: L. Default \(\leftrightarrow\) R. Default & + & 0.7208\\
Cryst & T+R & SPRN & 8 & FC Edge: L. Default \(\leftrightarrow\) R. Dorsal Attention & + & 0.7010\\
\end{longtable}
\endgroup

\renewcommand{\thetable}{S7d}
\begingroup\footnotesize
\begin{longtable}{@{\extracolsep{\fill}}lllllll@{}}
\caption{Top intrinsic resting-state (rsfMRI) functional connections restricted strictly to the top 10 highest-ranked nodes.}\\
\toprule
Target & Cohort & Model & Rank & Connection & Sign & Score\\\midrule\endfirsthead
\multicolumn{7}{l}{\tablename\ \thetable\ (continued)}\\\toprule
Target & Cohort & Model & Rank & Connection & Sign & Score\\\midrule\endhead\bottomrule\endfoot
Cryst & T+R & SPRN & 1 & L. Default \(\leftrightarrow\) R. Default & + & 0.3931\\
Cryst & T+R & SPRN & 2 & L. Ventral Attention \(\leftrightarrow\) R. Dorsal Attention & - & 0.3831\\
Cryst & T+R & SPRN & 3 & L. Ventral Attention \(\leftrightarrow\) L. Dorsal Attention & - & 0.3343\\
Cryst & T+R & SPRN & 4 & R. Dorsal Attention \(\leftrightarrow\) R. Default & + & 0.1868\\
Cryst & T+R & SPRN & 5 & L. Dorsal Attention \(\leftrightarrow\) R. Dorsal Attention & - & 0.1811\\
Cryst & T+R & SPRN & 6 & L. Ventral Attention \(\leftrightarrow\) L. Visual & - & 0.1255\\
Cryst & T+R & SPRN & 7 & L. Cingulo-Parietal \(\leftrightarrow\) L. Cingulo-Parietal & + & 0.1126\\
Cryst & T+R & SPRN & 8 & L. Cingulo-Parietal \(\leftrightarrow\) L. Ventral Diencephalon & + & 0.0880\\
\end{longtable}
\endgroup

\renewcommand{\thetable}{S8a}
\begingroup\footnotesize
\begin{longtable}{@{\extracolsep{\fill}}lllllll@{}}
\caption{Top 10 functional task (tfMRI) nodes, aggregated across all conditions and contrasts.}\\
\toprule
Target & Cohort & Model & Rank & Node & Sign & Score\\\midrule\endfirsthead
\multicolumn{7}{l}{\tablename\ \thetable\ (continued)}\\\toprule
Target & Cohort & Model & Rank & Node & Sign & Score\\\midrule\endhead\bottomrule\endfoot
Cryst & T+R & MACNET & 1 & R. Precentral & - & 4.4897\\
Cryst & T+R & MACNET & 2 & L. Pars Opercularis & - & 3.3357\\
Cryst & T+R & MACNET & 3 & L. Superior Parietal & - & 3.2978\\
Cryst & T+R & MACNET & 4 & L. Precentral & - & 3.2268\\
Cryst & T+R & MACNET & 5 & L. Lingual & + & 3.1179\\
Cryst & T+R & MACNET & 6 & L. Banks Superior Temporal Sulcus & - & 2.9984\\
Cryst & T+R & MACNET & 7 & R. Caudal Anterior Cingulate & + & 2.9238\\
Cryst & T+R & MACNET & 8 & R. Rostral Middle Frontal & + & 2.8934\\
\end{longtable}
\endgroup

\renewcommand{\thetable}{S8b}
\begingroup\footnotesize
\begin{longtable}{@{\extracolsep{\fill}}lllllll@{}}
\caption{Top 10 specific functional task (tfMRI) features, defined by distinct region-contrast pairings.}\\
\toprule
Target & Cohort & Model & Rank & Feature & Sign & Score\\\midrule\endfirsthead
\multicolumn{7}{l}{\tablename\ \thetable\ (continued)}\\\toprule
Target & Cohort & Model & Rank & Feature & Sign & Score\\\midrule\endhead\bottomrule\endfoot
Cryst & T+R & MACNET & 1 & R. Precentral [MID: Ant. Loss vs Neut] & - & 0.9530\\
Cryst & T+R & MACNET & 2 & R. Precentral [MID: Ant. Small Loss vs Neut] & - & 0.9477\\
Cryst & T+R & MACNET & 3 & R. Pars Triangularis [N-back: 0-back] & - & 0.6803\\
Cryst & T+R & MACNET & 4 & R. Caudal Anterior Cingulate [N-back: 2-back vs 0-back] & + & 0.6771\\
Cryst & T+R & MACNET & 5 & L. Pars Opercularis [N-back: 2-back] & - & 0.6415\\
Cryst & T+R & MACNET & 6 & L. Superior Parietal [N-back: 2-back] & - & 0.6284\\
Cryst & T+R & MACNET & 7 & L. Isthmus Cingulate [N-back: 0-back] & - & 0.6009\\
Cryst & T+R & MACNET & 8 & R. Precentral [MID: Ant. Large Loss vs Neut] & - & 0.5897\\
\end{longtable}
\endgroup

\vspace{1em}

\begin{center}
\textbf{List of Abbreviations}
\end{center}

\begingroup\footnotesize
\begin{longtable}{@{\extracolsep{\fill}}lp{12cm}@{}}
\toprule
Abbreviation & Definition \\\midrule\endfirsthead
\toprule
Abbreviation & Definition \\\midrule\endhead
\bottomrule\endfoot
ABCD & Adolescent Brain Cognitive Development \\
ASEG & Automated Segmentation (FreeSurfer parcellation) \\
BMI & Body Mass Index \\
C & Common-cohort-only training \\
DK & Desikan-Killiany (Cortical parcellation) \\
FA & Fractional Anisotropy \\
FC & Functional Connectivity \\
\(G_c\) & Crystallized Cognition Composite \\
\(G_f\) & Fluid Cognition Composite \\
\(G_t\) & Total Cognition Composite \\
IG & Integrated Gradients \\
IXG & Input \(\times\) Gradient \\
MAC-Net & Multi-task Activation and Contrast Network \\
MAE & Mean Absolute Error \\
MD & Mean Diffusivity \\
MID & Monetary Incentive Delay \\
MLP & Multilayer Perceptron \\
MRI & Magnetic Resonance Imaging \\
MSE & Mean Squared Error \\
\(R^2\) & Coefficient of Determination \\
RF & Random Forest \\
ROI & Region of Interest \\
rsfMRI & Resting-State Functional MRI \\
SAL & Saliency \\
sMRI & Structural and Diffusion MRI \\
SPRN & Sparse Projected Residual Network \\
tfMRI & Task-based Functional MRI \\
TS & Two-Stage opportunistic training \\
U & Unimodal \\
\end{longtable}
\endgroup

\end{document}